\pdfoutput=1

\documentclass[11pt]{article}
\usepackage[dvipsnames]{xcolor}

\usepackage{EMNLP2022}

\usepackage{times}
\usepackage{latexsym}

\usepackage[T1]{fontenc}

\usepackage[utf8]{inputenc}

\usepackage{microtype}

\usepackage{inconsolata}

\usepackage{times}
\usepackage{array}
\usepackage{arydshln}

\usepackage[flushleft]{threeparttable}

\usepackage{hyperref}
\usepackage{tabularx}
\usepackage{multirow}
\usepackage{enumitem}
\usepackage{fancyvrb}
\usepackage{booktabs} 
\usepackage{tikz}
\usetikzlibrary{patterns}
\usetikzlibrary{positioning}
\usepackage{diagbox}

\usepackage{graphicx}
\usepackage{caption}
\usepackage{subcaption}
\usepackage{bm}
\usepackage{url}
\usepackage{rotating}
\usepackage{pdflscape}
\usepackage{comment}
\usepackage{amssymb}
\usepackage{stfloats}
\usepackage{tabu}
\usepackage{bbm}
\usepackage[normalem]{ulem}

 \usepackage{pifont}
 
\usepackage{caption}

 \usepackage{float}
 
\usepackage{makecell}
 
\usepackage{soul}

\usepackage{amsmath}
\usepackage[export]{adjustbox}

\definecolor{colorAdd}{HTML}{ffd6d1}
\definecolor{colorSynonym}{HTML}{ffffd7}
\definecolor{colorPos}{HTML}{d4ebff}
\definecolor{colorStructure}{HTML}{fddced}
\definecolor{colorSimplification}{HTML}{feddb7}
\definecolor{colorCopy}{HTML}{e7e7e7}
\definecolor{colorSimple}{HTML}{e6ffcc}
\definecolor{colorHallu}{HTML}{dfdbfd}
\definecolor{colorMiss}{HTML}{ddf0bb}
\definecolor{colorMisinterpret}{HTML}{fad5df}
\definecolor{colorGrammar}{HTML}{cfffe8}

\definecolor{someSpanText}{HTML}{000000} 

\newcommand{\paraAdd[1]}{\sethlcolor{colorAdd}{\textbf{\hl{~#1~}}}}
\newcommand{\paraSynonym[1]}{{\sethlcolor{colorSynonym}\textcolor{someSpanText}{\textbf{\hl{~#1~}}}}}

\newcommand{\paraPos[1]}{\sethlcolor{colorPos}{\textbf{\hl{~#1~}}}}
\newcommand{\paraStructure[1]}{\sethlcolor{colorStructure}{\textcolor{someSpanText}{\textbf{\hl{~#1~}}}}}
\newcommand{\paraSimplification[1]}{\sethlcolor{colorSimplification}{\textcolor{someSpanText}{\textbf{\hl{~#1~}}}}}
\newcommand{\paraCopy[1]}{\sethlcolor{colorCopy}{\textcolor{someSpanText}{\textbf{\hl{~#1~}}}}}
\newcommand{\paraSimple[1]}{\sethlcolor{colorSimple}{\textcolor{someSpanText}{\textbf{\hl{~#1~}}}}}
\newcommand{\paraHallu[1]}{\sethlcolor{colorHallu}{\textbf{\hl{~#1~}}}}
\newcommand{\paraMiss[1]}{\sethlcolor{colorMiss}{\textcolor{someSpanText}{\textbf{\hl{~#1~}}}}}
\newcommand{\paraMisinterpret[1]}{\sethlcolor{colorMisinterpret}{\textcolor{someSpanText}{\textbf{\hl{~#1~}}}}}
\newcommand{\paraGrammar[1]}{\sethlcolor{colorGrammar}{\textcolor{someSpanText}{\textbf{\hl{~#1~}}}}}

\newcommand{\nameAdd}{\paraAdd[Add New]}
\newcommand{\nameSynonym}{\paraSynonym[Word Syn]}
\newcommand{\namePos}{\paraPos[Phrase Syn]}
\newcommand{\nameStructure}{\paraStructure[Structure]}
\newcommand{\nameSimplification}{\paraSimplification[Simplification]}
\newcommand{\nameCopy}{\paraCopy[Copy]}
\newcommand{\nameSimple}{\paraSimple[Small Change]}
\newcommand{\nameHallu}{\paraHallu[Hallucination]}
\newcommand{\nameMiss}{\paraMiss[Miss Info]}
\newcommand{\nameMisinterpret}{\paraMisinterpret[Misinterpret]}
\newcommand{\nameGrammar}{\paraGrammar[Bad Grammar]}

\newcommand{\datasetname}{\textsc{MultiPIT}}
\newcommand{\crowddata}{\datasetname\textsubscript{\textsc{crowd}}}
\newcommand{\expertdata}{\datasetname\textsubscript{\textsc{expert}}}
\newcommand{\newdata}{\datasetname\textsubscript{\textsc{Auto}}}
\newcommand{\newdataAbb}{\textsc{M}\textsubscript{\textsc{Auto}}}
\newcommand{\newtestdata}{\datasetname\textsubscript{\textsc{NMR}}}
\DeclareMathSymbol{\mh}{\mathord}{operators}{`\-}

\newlength\replength
\newcommand\repfrac{.33}

\newcommand\rulewidth{.6pt}
\newcommand\tdashfill[1][\repfrac]{\cleaders\hbox to \replength{%
  \smash{\rule[\arraystretch\ht\strutbox]{\repfrac\replength}{\rulewidth}}}\hfill}

\newcommand\tdotfill[1][\repfrac]{\cleaders\hbox to \replength{%
  \smash{\raisebox{\arraystretch\dimexpr\ht\strutbox-.1ex\relax}{.}}}\hfill}

\usepackage{array, booktabs, makecell, multirow}

\newcommand{\PreserveBackslash}[1]{\let\temp=\\#1\let\\=\temp}
\newcolumntype{C}[1]{>{\PreserveBackslash\centering}p{#1}}
\newcolumntype{R}[1]{>{\PreserveBackslash\raggedleft}p{#1}}
\newcolumntype{L}[1]{>{\PreserveBackslash\raggedright}p{#1}}

\title{Improving Large-scale Paraphrase Acquisition and Generation}

\newcommand{\changeurlcolor}[1]{\hypersetup{urlcolor=#1}}  

\author{Yao Dou, Chao Jiang, Wei Xu \\
  School of Interactive Computing \\
  Georgia Institute of Technology \\
 \texttt{\{douy, chaojiang\}@gatech.edu; wei.xu@cc.gatech.edu}
  \\
\changeurlcolor{orange}\url{http://twitter-paraphrase.com/}
\\}
\date{}

\begin{document}
\maketitle

\begin{abstract}


This paper addresses the quality issues in existing Twitter-based paraphrase datasets, and discusses the necessity of using two separate definitions of paraphrase for identification and generation tasks. We present a new Multi-Topic Paraphrase in Twitter (\datasetname) corpus that consists of a total of 130k sentence pairs with crowdsoursing (\crowddata) and expert (\expertdata) annotations using two different paraphrase definitions for paraphrase identification, in addition to a multi-reference test set (\newtestdata) and a large automatically constructed training set (\newdata) for paraphrase generation. With improved data annotation quality and task-specific paraphrase definition, the best pre-trained language model fine-tuned on our dataset achieves the state-of-the-art performance of 84.2 $F_1$ for automatic paraphrase identification. Furthermore, our empirical results also demonstrate that  the paraphrase generation models trained on \newdata~generate more diverse and high-quality paraphrases compared to their counterparts fine-tuned on other corpora such as Quora, MSCOCO, and ParaNMT.

\end{abstract}

\section{Introduction}

\begin{figure}[t!]

\begin{center}
\includegraphics[width=0.98\linewidth]{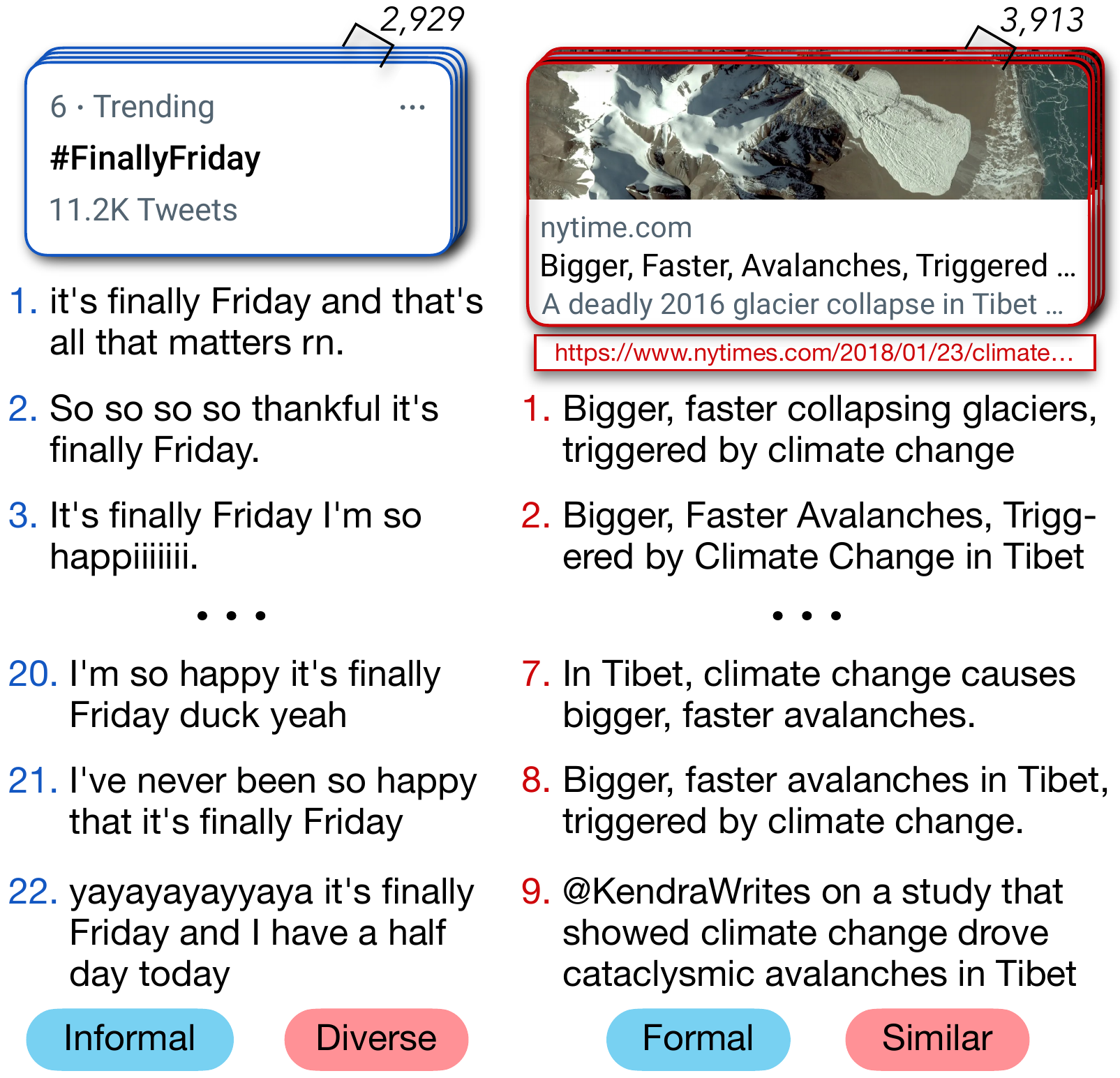}
\end{center}
\vspace{-9pt}
\caption{
Two sets of paraphrases in \datasetname, discussing a trending topic or a news article, respectively. 
}
\vspace{-15pt}
\label{fig:figure1}
\end{figure}




Paraphrases are alternative expressions that convey a similar meaning \cite{bhagat2013paraphrase}. 
Studying paraphrase  facilitates research in both natural language understanding and  generation.
For instance, identifying paraphrases on social media is important for tracking the spread of misinformation \cite{bakshy2011everyone} and capturing emerging events \cite{vosoughi2016semi}. 
On the other hand, paraphrase generation improves the linguistic diversity in conventional agents \cite{li-etal-2016-diversity} and machine translation \cite{thompson-post-2020-automatic}. It has also been successfully applied in data argumentation to improve information extraction \cite{zhang-etal-2015-exploiting, ferguson-etal-2018-semi} and question answering systems \cite{gan-ng-2019-improving}.
Many researchers have been leveraging Twitter data to study paraphrase given its lexical and style diversity as well as coverage of up-to-date events.
However, existing Twitter-based paraphrase datasets, namely PIT-2015 \cite{xu2015semeval} and Twitter-URL \cite{lan-etal-2017-continuously}, suffer from quality issues such as topic unbalance and annotation noise,\footnote{63\% of sentences in Twitter-URL are related to the 2016 US presidential election, and  58\% of sentences in PIT-2015 are about NFL draft  (more detailed analysis in $\S$ \ref{section:quality_control}).} which limit the performance of the models trained using them. Moreover, past efforts on creating paraphrase corpora only consider one paraphrase criteria without taking into account the fact that the desired ``strictness'' of semantic equivalence in paraphrases varies from task to task \cite{bhagat2013paraphrase,liu-soh-2022-towards}. For example, for the purpose of tracking unfolding events, \textit{``A tsunami hit Haiti.''} and \textit{``303 people died because of the tsunami in Haiti''} are sufficiently close to be considered as paraphrases; whereas for paraphrase generation, the extra information \textit{``303 people dead''} in the latter sentence may lead models to learn to hallucinate and generate more unfaithful content. 

\setlength{\tabcolsep}{3pt}
\begin{table*}[hpt!]
\centering
\resizebox{\textwidth}{!}{
\renewcommand{\arraystretch}{1}
\begin{tabular}{p{.15in}lrrrrrrrr} 
\toprule
  \multicolumn{2}{l}{\textbf{Topic Domains}}  & \textbf{\#Train} &  \textbf{\#Dev} & \textbf{\#Test} & \textbf{Sent/Tweet Len} & \textbf{\%Paraphrase} & \textbf{\#Trends/URLs}  & \textbf{\#Uniq Sent} & \textbf{\%Multi-Ref}  \\ 
  \midrule
  \multicolumn{9}{l}{\textit{\textbf{Our Multi-Topic Paraphrase in Twitter  (\crowddata) Dataset}}}                          \\
\multirow{4}{*}{\rotatebox[origin=c]{90}{Trends}} & Sports                  & 25,255  & 3,157 &  3,157 & 10.24 / 13.79 & 40.52\% & 1,201 & 34,786 & 17.89\% \\ 
& Entertainment    &  11,547 & 1,443 & 1,444  & 10.44 / 13.80 & 62.33\% & 610 & 15,784 & 18.11\% \\ 
& Event   & 8,624 & 1,078 & 1,079 & 10.86 / 15.32 & 82.83\% & 359 & 11,746 & 17.75\% \\ 
& Others  & 17,751  & 2,219  &  2,219 & 10.41  / 14.56 & 67.16\% & 817 & 24,286 & 18.33\% \\ 
\midrule

\multirow{4}{*}{\rotatebox[origin=c]{90}{URL}} &  Science/Tech          & 7,384  & 923 & 923  & 10.94 / 19.17 & 46.13\% & 1,032 & 10,327 & 17.74\% \\ 

& Health & 9,123  & 1,140  & 1,141 & 11.29 / 21.68 & 46.78\% & 1,298 & 12,772 & 17.86\% \\ 
& Politics   & 7,981   & 998  & 998 & 10.95 /  18.48 & 56.56\% & 1,063 & 10,999 & 17.68\% \\ 

& Finance     &  4,552  & 569    & 569   & 11.19 / 23.08 & 18.96\% & 554 & 5,907 & 20.13\% \\ 
\multicolumn{2}{l}{\enskip \textbf{Total}} & \textbf{92,217}   & \textbf{11,527}   & \textbf{11,530}  & \textbf{10.62}  / \textbf{16.10}  & \textbf{53.73\%}  & \textbf{6,934}  & \textbf{124,438}  & \textbf{18.65\%} \\ 


\midrule

\multicolumn{2}{l}{\textit{\textbf{Our \expertdata~Dataset}}}  &    4,458 &  555 & 557 & 12.08 / 17.02 & 53.11\% & 200 & 5,743 & 100\% \\ 

\midrule
\midrule

  \multicolumn{9}{l}{\textit{{Existing Twitter Paraphrase Datasets}}}                          \\ 
\multicolumn{2}{l}{PIT-2015 (\citeauthor{xu2015semeval})}        & 13,063    & 4,727        & 972     & 11.9 / $\,$ -- $\,$   & 30.60\% & 420   & 19,297       & 24.67\%       \\ 
\multicolumn{2}{l}{Twitter URL (\citeauthor{lan-etal-2017-continuously})}       & 42,200    & $\;$ -- $\; \;$        & 9,324   & $\,$ -- $\,$ / 14.8           & 22.77\% &      5,187 & 48,906 &23.91\%    \\ \bottomrule
\end{tabular}}
\vspace{-6pt}
\caption{Statistics of \crowddata~and \expertdata~datasets. The sentence/tweet lengths are calculated based on the number of tokens per unique sentence/tweet. \%Multi-Ref denotes the percentage of source sentences with more than one paraphrase. Compared with prior work, our \crowddata~dataset has a significantly larger size, a higher portion of paraphrases, and a more balanced topic distribution.}
\vspace{-8pt}
\label{table:statistics_on_datasets}
\end{table*}

In this paper, we present an effective data collection and annotation method to address these issues. We curate the Multi-Topic Paraphrase in Twitter (\datasetname) corpus, which includes \crowddata, a large crowdsourced set of 125K sentence pairs that is useful for tracking information on Twitter, and \expertdata, an expert annotated set of 5.5K sentence pairs using a stricter definition that is more suitable for acquiring paraphrases for generation purpose. Compared to PIT-2015 and Twitter-URL, our corpus contains more than twice as much data with more balanced topic distribution and better annotation quality. Two sets of examples from \datasetname~are shown in Figure \ref{fig:figure1}.

We extensively evaluate several state-of-the-art neural language models on our datasets to demonstrate the importance of having task-specific paraphrase definition. Our best model achieves 84.2 F$_1$ for automatic paraphrase identification. In addition, we construct a continually growing paraphrase dataset, \newdata, by applying the automatic identification model to unlabelled Twitter data. Empirical results and analysis show that generation models fine-tuned on \newdata~generate more diverse and high-quality paraphrases compared to models trained on other corpora, such as MSCOCO \cite{lin2014microsoft},  ParaNMT \cite{wieting-gimpel-2018-paranmt}, and Quora.\footnote{\url{https://www.kaggle.com/c/quora-question-pairs}} We hope our \datasetname~corpus will facilitate future innovation in paraphrase research.

\section{Multi-Topic PIT Corpus}
\label{sec:dataset}


In this section, we present our data collection and annotation methodology for creating \crowddata~and \expertdata~datasets. The data statistics is detailed in Table \ref{table:statistics_on_datasets}.
\subsection{Collection of Tweets}
\label{subsec:tweets_grouping}
To gather paraphrases about a diverse set of topics as illustrated in Figure \ref{fig:figure1}, we first group tweets that  contain the same trending topic\footnote{\url{https://www.twitter.com/explore/tabs/trending}} (year 2014--2015) or the same URL (year 2017--2019) retrieved through Twitter public APIs\footnote{\url{https://developer.twitter.com/en/docs/twitter-api}} over a long time period. 
Specifically, for the URL-based method, we extract the URLs embedded in the tweets that are posted by 15 news agency accounts (e.g., \textit{NYTScience}, \textit{CNNPolitics}, and \textit{ForbesTech}). To get cleaner paraphrases, we split the tweets into sentences, eliminating the extra noises caused by multi-sentence tweets. 
More details of the improvements we  made to address the  data preprocessing  issues in prior work are described in Appendix \ref{appendix:preprocessing}.





\begin{figure*}
\centering
\begin{subfigure}{.33\textwidth}
  \centering
  \includegraphics[width=\linewidth]{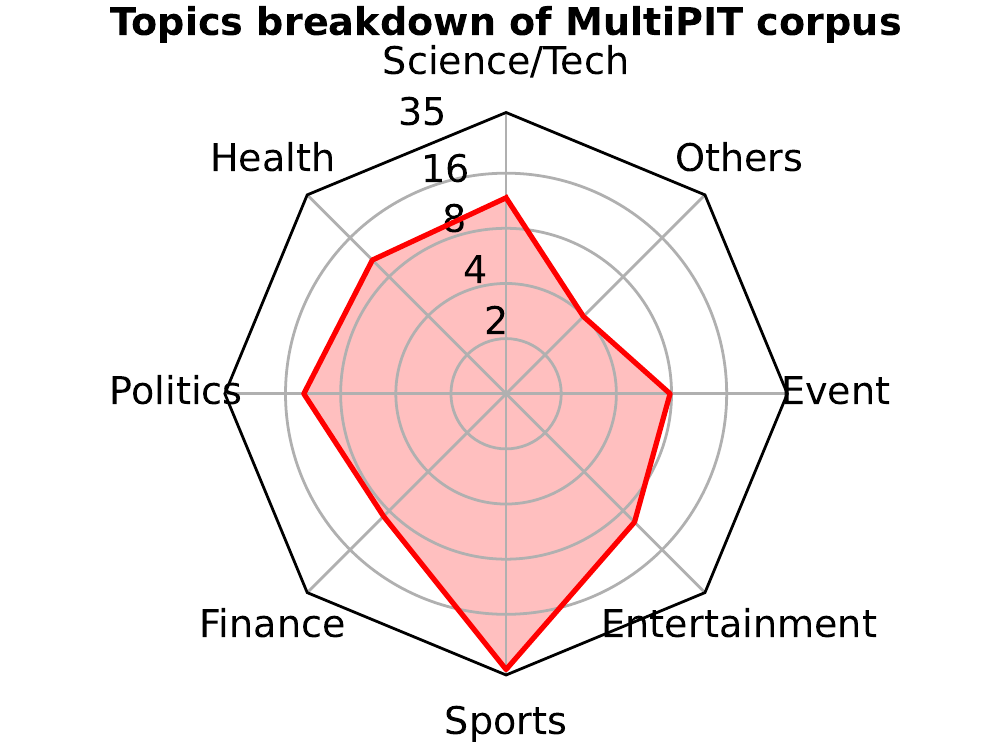}
  \label{fig:sfig1}
\end{subfigure}%
\begin{subfigure}{.33\textwidth}
  \centering
  \includegraphics[width=\linewidth]{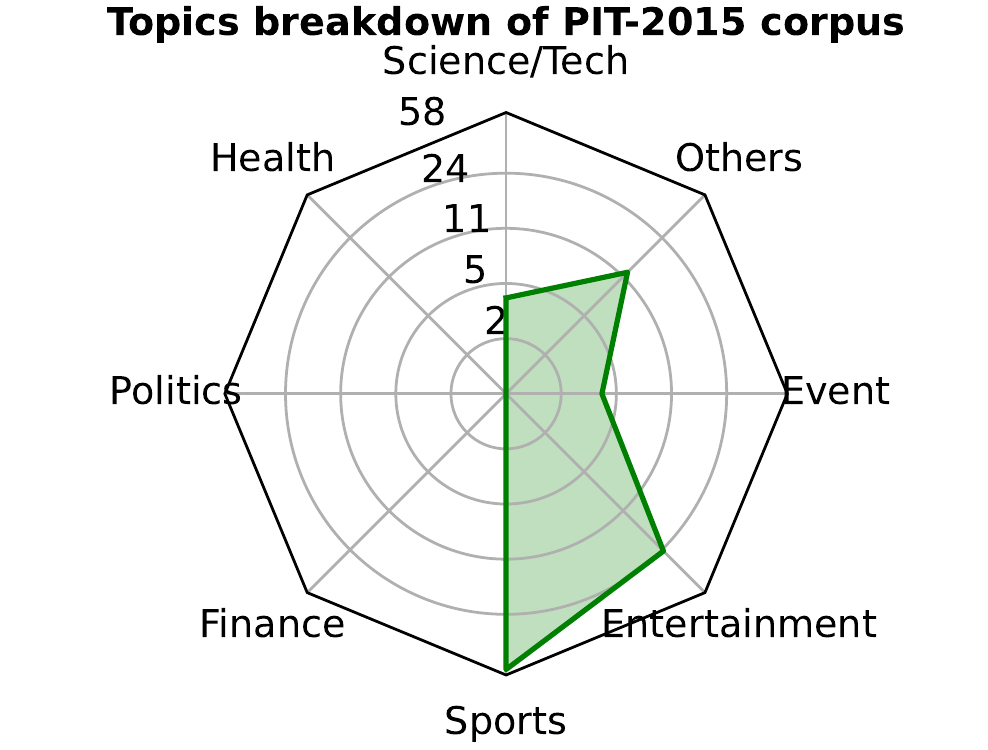}
  \label{fig:sfig2}
\end{subfigure}
\begin{subfigure}{.33\textwidth}
  \centering
  \includegraphics[width=\linewidth]{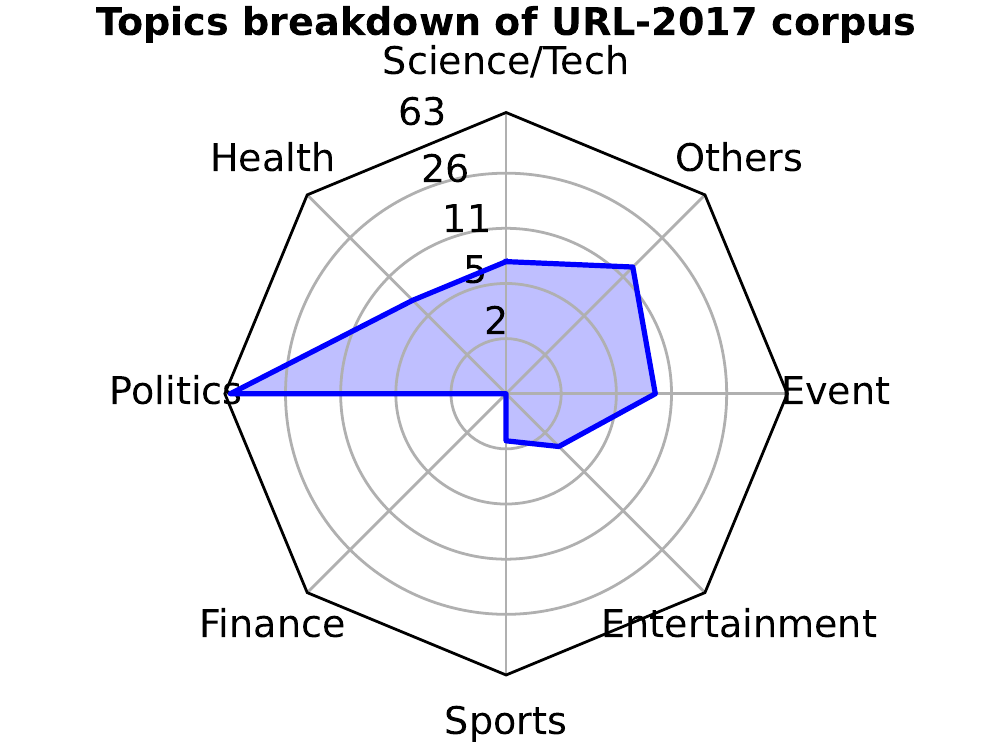}
  \label{fig:sfig3}
\end{subfigure}
\vspace{-19pt}
\caption{
Topic breakdown on 100 randomly sampled sentence pairs from \crowddata, PIT-2015 and Twitter-URL. Our \crowddata~corpus has a more balanced topic distribution.
}
\vspace{-8pt}
\label{fig:topics_breakdown}
\end{figure*}

\subsection{Topic Classification and Balancing}
\label{subsec:topic_classification}

To avoid a single type of topics dominating the entire dataset as in prior work \cite{xu2015semeval,lan-etal-2017-continuously}, we manually categorize the topics for each group of tweets and balance their distribution.  
For trending topics, we ask three in-house annotators to classify them into 4 different categories: sports, entertainment, event, and others. 
All three annotators are college students with varied linguistic annotation experience, and each received an hour-long training session. 
For URLs, most of them are linked to news articles and have already been categorized by the news agency.\footnote{For example, URL \url{https://www.nytimes.com/2019/08/09/science/komodo-dragon-genome.html} belongs to science topic.} 
We include the tweets grouped by URLs that belong to the science/tech, health, politics, and finance categories.

\subsection{Candidate Selection}
The PIT-2015 \cite{xu2015semeval} and Twitter-URL \cite{lan-etal-2017-continuously} corpora contain only 23\% and 31\% sentence pairs that are paraphrases, respectively. To increase the portion of paraphrases and improve the annotation efficiency, we introduce an additional step to filter out the tweet groups that contain either too much noise or too few paraphrases, and adaptively select sentence pairs for  annotation (\S \ref{section:quality_control}).
For each of the trend-based groups, we first select the top 2 sentences using a simple ranking algorithm \cite{xu2015semeval} based on the averaged probability of words. 
We pair each of these two sentences with 10 other sentences that are randomly sampled from the top 20 in each group. 
Among these 20 sentence pairs, if the annotators found $n \in$ [4, 6] or [7, 9] or [10, 12] or [13, 20] pairs as paraphrases, then we further deploy 20, 30, 40, or 50 sentence pairs for annotation, respectively. We pair one of the top 5 ranked sentences with 10 sentences randomly selected from those ranked between top 6 and top 50. 
Since the URL-based groups  generally contain fewer sentences, we select the top 11 sentences and ask annotators to choose one as the seed sentence that can be paired with the rest 10 sentences to produce at least 3 paraphrase pairs. 
If such a seed sentence exists, we pair it with the rest 10 sentences and  deploy them for annotation. Otherwise, we skip the entire group.  

\subsection{Crowd Annotation for Paraphrase Identification}
\label{section:quality_control}

We then annotate the selected sentence pairs using the crowdsourcing platform  Figure-Eight\footnote{\url{https://www.appen.com/}} to construct  \crowddata.



\begin{figure*}[t!]

\begin{center}
\includegraphics[width=0.99\linewidth]{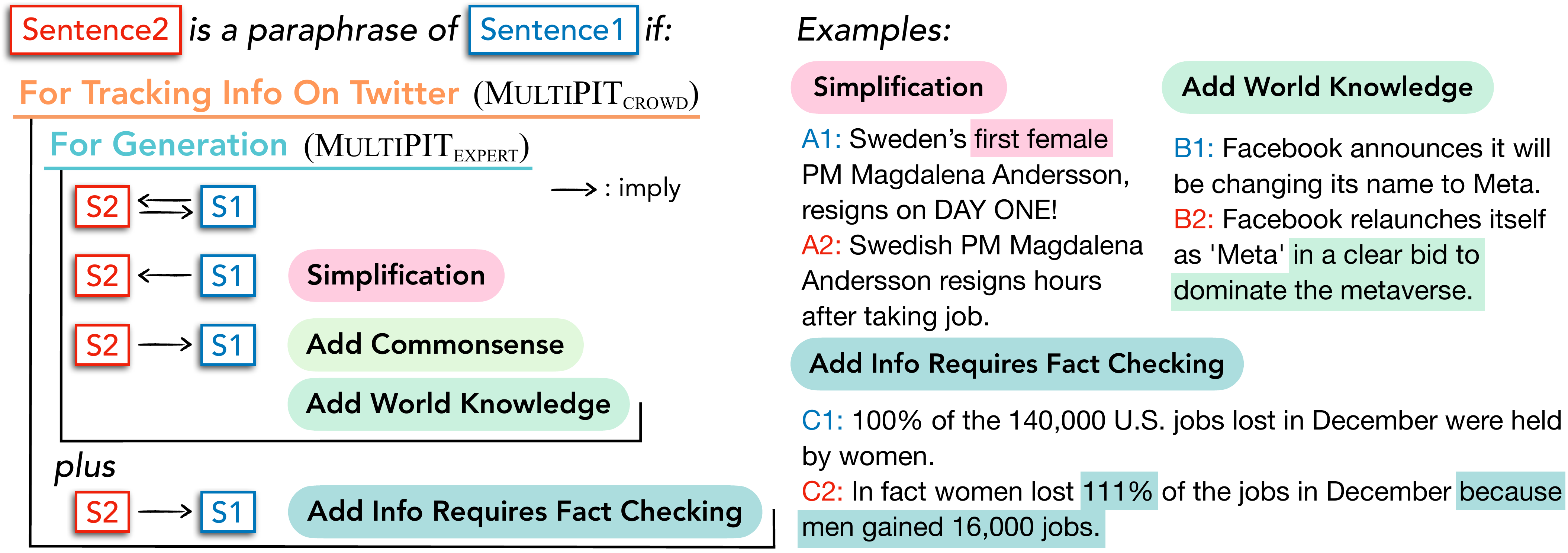}
\end{center}
\vspace{-6pt}
\caption{
Two different paraphrase definitions used for creating \crowddata~and \expertdata, with examples. 
The difference between the two criteria is whether considering \textit{Sentence2} that contains new information that requires fact-checking as a paraphrase of \textit{Sentence1}.\footnotemark}
\vspace{-8pt}
\label{fig:identification_vs_generation}
\end{figure*}

\paragraph{Annotation Process.} 
We design a 1-vs-1 annotation schema, where we present one sentence pair to workers at a time and ask them to annotate whether it is a paraphrase pair or not. A screenshot of the annotation interface is provided in Appendix \ref{appendix:crowd-annotation}. We collect 6 judgments for every sentence pair and pay \$0.2 per annotation ($>$\$7 per hour). For creating \crowddata, with the purpose of identifying similar sentences and tracking information spreading on Twitter in mind, we consider two sentences as paraphrases even if one contains some new information that does not appear in the other sentence (see Figure \ref{fig:identification_vs_generation} for examples). As a side note, because these sentences are grouped under the same trend or URL, the new information is always relevant and based on the context, otherwise, we will consider them non-paraphrases.

\paragraph{Quality Control.}
In every five sentence pairs, we embed one hidden test sentence pair that are pre-labeled by one of the authors, and constantly monitor the workers' performance.
Whenever annotators make a mistake on the test pair, they will be alerted and provided with an explanation. Workers can continue in the task if they achieve  $>$85\% accuracy on the test pairs and $>$0.2 Cohen's \cite{Cohen1960ACO} kappa when compared with the major vote of other workers. All workers are  in the U.S.

\paragraph{Inter-Annotator Agreement.} The average Cohen's kappa is 0.75 for URL-sourced sentence pairs, 0.69 for Trends-sourced ones, and 0.70 for all. 
We also sample 400 sampled sentence pairs and hire two experienced in-house annotators to label them. Assuming the in-house annotation is gold, the $F_1$ of crowdworkers' majority vote is 89.1.

\paragraph{Accessing Topic Diversity.}
We manually examine 100 sentence pairs randomly sampled  from \crowddata, PIT-2015 \cite{xu2015semeval} and Twitter-URL \cite{lan-etal-2017-continuously}.
Figure \ref{fig:topics_breakdown} shows the results of the manual inspection. 
\crowddata~has a much more balanced topic distribution, compared to prior work where 58\% of sentences in PIT-2015 are about sports and 63\% of sentences in Twitter-URL are politics-related.
This improvement can be attributed to the long time periodd (\S \ref{subsec:tweets_grouping}) and topic classification step (\S \ref{subsec:topic_classification}) in our data collection process.
In contrast, PIT-2015 was collected within only 10 days  (04/24/2013 – 05/03/2013) that was overwhelmed by a popular sports event -- the 2013 NFL draft (04/25 - 04/27), and Twitter-URL was collected during the 3 months of the 2016 US presidential election.



\footnotetext{The example C1 and C2 is on the more extreme side of the "loose" paraphrase criterion from the linguistic perspective, more average cases are shown in Figure \ref{fig:figure1}.}

\subsection{Expert Annotation for Paraphrase Generation}
\label{subsec:multipit_expert}

Text generation models are prone to memorize training data and generate unfaithful hallucinations \cite{Maynez2020OnFA,Carlini2021ExtractingTD}. Including paraphrase pairs that contain extra information other than world or commonsense knowledge in the training data only worsens the problem, as shown in Table \ref{table:generation_variations} in Appendix \ref{sec:further_generation_experiment}. For the purpose of paraphrase generation, we further create \expertdata~with expert annotations, using a stricter paraphrase definition than the one used in \crowddata. The different paraphrase criteria used for creating these two datasets and their corresponding examples are illustrated in Figure \ref{fig:identification_vs_generation}.

\paragraph{Data Selection.} To create a high-quality corpus that focuses on differentiating strict paraphrases from the more loosely  defined ones, we first use our best paraphrase identifier (\S \ref{sec:identification}) fine-tuned on \crowddata~to filter the sentence pairs and then have experienced in-house annotators to further annotate them. Specifically, we gather sentence pairs that are identified as paraphrases by the automatic classifier from 9,762 trending topic groups (from Oct-Dec 2021) and 181,254 URL groups (from Jan 2020-Jun 2021). To improve the diversity of our dataset, instead of presenting these pairs directly to the experts for annotation, we cluster the sentences by considering the  paraphrase relationship transitive, i.e., if sentence pairs $(s_1, s_2)$ and $(s_2, s_3)$ are both identified as paraphrases, then $(s_1, s_2, s_3)$ is a cluster. For each trend or URL, we show two seed sentences paired with up to 30 sentences in the largest cluster for the experts to annotate. In total, we have 5,570 sentence pairs annotated for \expertdata, in which 100 sentences sourced by trend and 100 ones sourced by URL have at least 8 corresponding paraphrases. We use these 200 sets to form \newtestdata, the first multi-reference test set for paraphrase generation evaluation (\S \ref{sec:generation}).

\paragraph{Expert Annotation.}
We ask two experienced annotators with linguistic backgrounds and rich annotation experience to annotate each sentence pair as paraphrases or not.
Annotators thoroughly discuss pairs that have inconsistent judgments until reaching an agreement.
A screenshot of the updated annotation instruction is provided in Appendix \ref{appendix:expert-annotation}.
\section{Paraphrase Identification}
\label{sec:identification}

\begin{table*}[t]
\renewcommand{\arraystretch}{1}
\small
\centering
\resizebox{\linewidth}{!}{
\begin{tabular}{lccccccccccc}

\toprule
 \multirow{2}{*}[-4pt]{\textbf{Model}} & \multirow{2}{*}[-4pt]{\textbf{\#Para.}} &  \multicolumn{5}{c}{\textbf{\crowddata}} & \multicolumn{5}{c}{\textbf{\expertdata}}\\
\cmidrule(lr){3-7} \cmidrule(lr){8-12}
  & & LR & Precision & Recall & $F_1$ & Accuracy & LR & Precision & Recall & $F_1$ & Accuracy\\
  
   \midrule
   ESIM & 17M & 4e-4 & 89.55 & 70.15 & 78.67 & 82.15 & 4e-4 & 47.07 & 91.73 & 62.22 & 49.19 \\

    Infersent & 47M & 1e-3 & 87.03 & 87.57 & 87.29 & 86.47 & 1e-3 & 45.87 & \textbf{98.43} & 62.58 & 46.32 \\
    \midrule
    T5\textsubscript{base}   & 220M  & 1e-4 & 89.21 & 93.76 & 91.43 & 90.67 & 1e-4 & 71.96 & 83.86 & 77.45 & 77.74\\
    T5\textsubscript{large}  &   770M & 1e-5 & 90.36 & 93.58 & 91.94 & 91.29 & 1e-4 & 79.78 & 85.43 & 82.51 & 83.48 \\
    BERT\textsubscript{base}    & 109M & 3e-5 & 88.59 & 91.24 & 89.90 & 89.12 & 2e-5 & 71.66 & 86.61 & 78.43 & 78.28 \\
    BERT\textsubscript{large}    & 335M & 2e-5 & 88.73 & 93.17 & 90.90 & 90.10 & 2e-5 & 72.22 & 87.01 & 78.93 & 78.82  \\
    RoBERTa\textsubscript{large} &  355M & 2e-5 & \textbf{90.81} & 92.70 & 91.74 & 91.14 & 2e-5 & 77.01 & 83.07 & 79.92 & 80.97  \\
    BERTweet\textsubscript{large} & 355M & 2e-5 & 89.72 & \textbf{93.95} & 91.79 & 91.08 & 2e-5&  82.47 & 81.50 & 81.98 & 83.66 \\
    ALBERTV2\textsubscript{xxlarge}   & 235M & 1e-5 & 90.36 & 92.96 & 91.64 & 91.00 & 2e-5& \textbf{82.68} & 82.68 & 82.68 & 84.20 \\
    DeBERTaV3\textsubscript{large} & 400M & 5e-6 & 90.46 & 93.59 & \textbf{92.00} & \textbf{91.36} & 5e-6 & 82.56 & 83.86 & \textbf{83.20} & \textbf{84.56} \\
    \bottomrule
\end{tabular}
}
\caption{Results on the test sets of \crowddata~and \expertdata. Models are fine-tuned on the corresponding training set. DeBERTaV3\textsubscript{large} performs the best on both datasets. LR: learning rate.}
\vspace{-10pt}
\label{table:identification_main_v3}
\end{table*}

Paraphrase identification is a task that determines whether  two given sentences are paraphrases or not.
The two paraphrase definitions used in \crowddata~and \expertdata~suit different downstream applications: tracking information on Twitter and acquiring high-quality paraphrase pairs for training generation models.
Paraphrase identification models trained on our datasets achieve over 84 $F_1$ for each case.
\paragraph{Experimental Setup.} 
As each sentence pair in \crowddata~has six judgments, we use 3 as the threshold, where pairs with  $>$3 paraphrase judgments are labeled as paraphrase, and the ones with $<$3 paraphrase judgments are labeled as non-paraphrase.
We split \crowddata~and \expertdata~into 80/10/10\% for train/dev/test partitions by time such that the oldest data are used for training.
More details on the implementation and hyperparameter tuning are in Appendix \ref{sec:implementation_details}.

\subsection{Models}
We consider an encoder-decoder language model, T5 \cite{raffel2020exploring}, five masked language models, \textbf{BERT} \cite{devlin2018bert}, \textbf{RoBERTa} \cite{Liu2019RoBERTaAR}, \textbf{ALBERT} \cite{lan2019albert}, \textbf{BERTweet} \cite{nguyen-etal-2020-bertweet}, and \textbf{DeBERTaV3} \cite{He2021DeBERTaV3ID}. We also include  two competitive BiLSTM-based models, \textbf{Infersent} \cite{conneau-EtAl:2017:EMNLP2017} and \textbf{ESIM} \cite{chen2016enhanced}, to establish comparison with pre-BERT era work.

\subsection{Results}
Table \ref{table:identification_main_v3} presents   results for  the models fine-tuned on  each dataset.
DeBERTaV3\textsubscript{large} achieves the best results with 92 $F_1$ on \crowddata~and 83.2 $F_1$ on \expertdata.
Transformer-based models consistently outperform BiLSTM-based models, especially on \expertdata. 

\begin{table}[t]
\renewcommand{\arraystretch}{1}
\small
\centering
\resizebox{\linewidth}{!}{
\begin{tabular}{llcccc}

\toprule
Method & Data & P. & R. & $F_1$ & Acc. \\
    \midrule
    Fine-tuning & $\textsc{M}_\textsc{c}$ & 61.81 & \textbf{88.58} & 72.82 & 69.84 \\
    Fine-tuning & $\textsc{M}_\textsc{e}$ & 82.56 & 83.86 & 83.20 & 84.56 \\
    \midrule
    
    Fine-tuning & $\textsc{M}_\textsc{c}$ + $\textsc{M}_\textsc{e}$ & 62.99 & 87.80 & 73.36 & 70.92 \\
    + Filtering & $\textsc{M}_\textsc{c}$ + $\textsc{M}_\textsc{e}$ & 77.24 & 88.19 & 82.35 & 82.76 \\
    + Flipping & $\textsc{M}_\textsc{c}$ + $\textsc{M}_\textsc{e}$ & \textbf{83.40} & 85.04 & \textbf{84.21} & \textbf{85.46} \\
    \bottomrule
\end{tabular}
}
\caption{Results of different methods on the test set of \expertdata. $\textsc{M}_\textsc{c}$: \crowddata, $\textsc{M}_\textsc{e}$: \expertdata. We use DeBERTaV3\textsubscript{large} in the experiments.}
\vspace{-12pt}
\label{table:identification_variations}
\end{table}

\paragraph{Beyond Fine-tuning.} As \crowddata~is a large-scale dataset annotated with a loose paraphrase definition, we test whether leveraging these ``noisy" data improves model performance on \expertdata.
To reduce the noise  that comes from the difference in definitions, we first adjust the labeling threshold for \crowddata~from 3 to 4.
Then we consider two noisy training techniques adopted in  prior work \cite{Xie2020SelfTrainingWN, Zhang2018GeneralizedCE}, namely \textit{\textbf{filtering}}  and \textit{\textbf{flipping}}.
Specifically, we fine-tune a teacher model on \expertdata~and use it to go through \crowddata~as follows: for each sentence pair $p$, if its label is $i$ (0 for non-paraphrase, 1 for paraphrase) and $P_{\text{teacher}}(y=i|p) \leq \lambda$, we \textit{\textbf{filter}} out $p$  or \textit{\textbf{flip}} its label to $1-i$ (i.e. 0 $\rightarrow$ 1).\footnote{We perform a small grid search on $\lambda$ over  \{0.05, 0.15, 0.25, 0.35, 0.45\}, and find 0.35 works well for the \textit{filtering} method and 0.25 for the \textit{flipping} method.}
Next, we fine-tune a new model on the combination of \expertdata~ and the re-labeled \crowddata.
The experimental results are shown in Table \ref{table:identification_variations}.
Compared to fine-tuning on \expertdata, adding the original \crowddata~to the training data results in a 9.8 and 19.5  points drop in $F_1$ and  precision, respectively, demonstrating the necessity of task-specific paraphrase definition. 
Among all methods, the \textit{flipping} approach achieves the best $F_1$ of 84.2. We thus use it  to create \newdata~(\S\ref{sec:generation}). 

\subsection{Impact of Data Size}
\begin{figure}[t]

\begin{center}

\includegraphics[width=\linewidth]{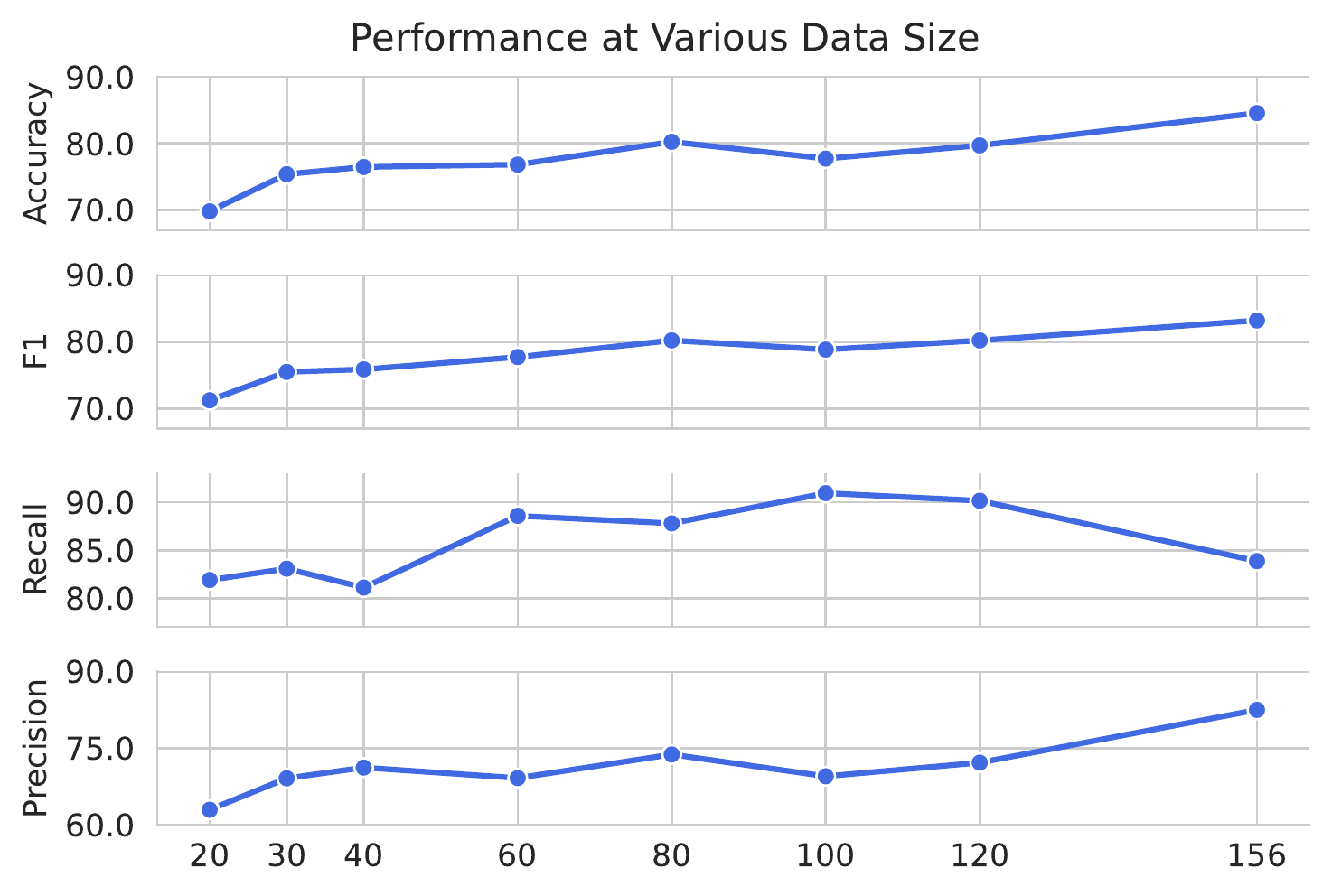}

\end{center}
\vspace{-10pt}
\caption{
Test set performance of model fine-tuned on varying amounts of data in \expertdata.
}
\vspace{-10pt}
\label{fig:datasize_vs_performance_identification}
\end{figure}

Figure \ref{fig:datasize_vs_performance_identification} shows test set performance of DeBERTaV3\textsubscript{large} fine-tuned on different amounts of data in \expertdata.
As there are 156 trend/URL groups in the train set, we truncate the data by group.
With more training data, the model achieves better $F_1$ and accuracy but in a slower fashion compared to the early stage.
This finding suggests that annotating more data can further improve the model's performance.

\section{Paraphrase Generation}
\label{sec:generation}

Paraphrase generation is a task that rewrites the input sentence while preserving its semantic meaning.
Since new  data is generated on Twitter every day, we introduce \newdata, an automated continual growing dataset for paraphrase generation.
We show that the model fine-tuned on \newdata~generates more diverse and high-quality paraphrases than other paraphrase datasets.

\subsection{Comparison with Existing Datasets}
\textbf{MSCOCO} \cite{lin2014microsoft}, and \textbf{ParaNMT} \cite{wieting-gimpel-2018-paranmt}, and \textbf{Quora}\footnote{\url{https://www.kaggle.com/c/quora-question-pairs}}  are three widely used datasets in paraphrase generation research \cite{zhou-bhat-2021-paraphrase}.
The Quora dataset contains over 400K question pairs, including 144K pairs labeled as duplicated (i.e., paraphrase), which are split into 134K/5K/5K as train/dev/test sets, respectively.
MSCOCO consists of over 120K images, each of which has five captions. 
Following \citet{Chen2020ASC}, for each image, we randomly pick  a caption and pair it with each of the other four  captions, resulting in about 490K paraphrase pairs. We split them into train/dev/test sets with 330K/80K/80K pairs, respectively.
ParaNMT is a dataset with more than 50 million paraphrase pairs that are automatically generated through back-translation. Since back-translation may introduce noise, we use the manually labeled dev and test sets from \citet{Chen2019ControllablePG}, which contain 499 and 871 instances, respectively.
\paragraph{\newdata.} We use the best performing model in Section \ref{sec:identification} to extract paraphrase pairs from recent Twitter data (trending topics in Oct-Dec 2021 and URLs in Jan 2020-Jun 2021).
We call these automated identified paraphrase pairs \newdata,\footnote{Future identified paraphrase pairs will be released every month.} which contains 302,307 pairs.
One of the authors manually annotates 215 paraphrase pairs and uses them as the dev set.
We use the multi-reference \newtestdata~test set  (\S\ref{subsec:multipit_expert}) for evaluation.
As the test set and \newdata~come from the same time period, we filter out sentence pairs in \newdata~that share similar trends or URLs with the pairs from the test set. This leaves us with 290,395 pairs as the training set. 

Following \citet{Chen2019ControllablePG}, we remove paraphrase pairs with high BLEU scores  in each training set to ensure there is enough variation between paraphrases, leaving about 137K pairs for \newdata, 47K for Quora, 275K for MSCOCO, and 443K for ParaNMT.
Table \ref{table:bleu_filtering_datasets} in Appendix \ref{sec:further_generation_experiment} shows BLEU filtering improves model performance for all datasets.
Detailed dataset statistics are provided in Appendix \ref{sec:details:statistics}.

\subsection{Evaluation Metrics}
We consider four automated metrics that are commonly used in previous work \cite{li-etal-2019-decomposable, Niu2021UnsupervisedPW} for paraphrase generation: \textbf{BLEU} \cite{Papineni2002BleuAM}, \textbf{Self-BLEU} \cite{Liu2021OnLT}, \textbf{BERT-Score} \cite{Zhang2020BERTScoreET}, and \textbf{BERT-iBLEU} \cite{Niu2021UnsupervisedPW}. Self-BLEU is BLEU computed between the source sentence and the output, which measures surface-form diversity. BERT-Score is also calculated between the source sentence and the output, measuring semantic similarity. BERT-iBLEU is a harmonic mean of BERT-Score and $1-$Self-BLEU, encouraging both semantic similarity and diversity. We use SacreBLEU \cite{post-2018-call} to compute BLEU and \textit{bert-score}\footnote{\url{https://www.github.com/Tiiiger/bert_score} We use DeBERTa-xlarge-mnli since it has the best correlation with human evaluation according  to \citet{Zhang2020BERTScoreET}.} to compute BERT-Score.

\begin{table}[t]
\centering
\renewcommand{\arraystretch}{1}
\resizebox{0.95\linewidth}{!}{
\begin{tabular}{@{}lcccccc@{}}
\toprule
   Model & \#Para. & LR & BL & S-B $\downarrow$ & B-S & B-iB  \\ 
   \midrule
      GPT-2\textsubscript{small}   & 117M & 3e-5 &  41.15 & 51.38 & 88.18 & 65.23  \\ 
    GPT-2\textsubscript{large}   & 774M  & 3e-5 & 42.89 & 39.61 & 86.16 & 74.01 \\ 
   BART\textsubscript{base}  & 139M  & 1e-5 &  46.91 & 46.38 & 87.65 & 71.40  \\ 
    BART\textsubscript{large} & 406M & 1e-5 & \textbf{47.22} & 38.26 & 86.40 & 75.17 \\ 
    T5\textsubscript{small} & 60M  & 3e-4 & 38.27 & 52.16 & \textbf{88.32} & 68.37  \\ 
    T5\textsubscript{base}  & 220M  & 1e-4 & 42.10 & 46.43 & 87.75 & 72.29 \\ 
    T5\textsubscript{large} & 770M & 1e-4 & 41.14 & \textbf{33.34} & 85.86 & \textbf{77.79} \\ 
    \midrule
    GPT-3\textsubscript{zero-shot} & 175B & - & 28.05 & 31.68 & 86.66 & 80.16\\
    GPT-3\textsubscript{few-shot} & 175B & - & 30.17 & 30.93 & 86.84 & 81.13\\
    \midrule
    \midrule
     \textit{Diversity (S-B $\downarrow$)}   & \multicolumn{2}{c}{\textit{Min.}} & \multicolumn{2}{c}{\textit{Avg.}} & \multicolumn{2}{c}{\textit{Max.}} \\ 
     Human Reference& \multicolumn{2}{c}{6.52} & \multicolumn{2}{c}{17.45\textsubscript{$\pm$9.14}} & \multicolumn{2}{c}{34.06}\\
    \bottomrule
\end{tabular}}
\caption{Test set results of different transformer models fine-tuned on \newdata, except GPT-3, where in-context learning is used. BL: BLEU, S-B: Self-BLEU, B-S: BERT-Score, B-iB: BERT-iBLEU. LR: learning rate. \textbf{Bold}: the best. The Self-BLEU of human reference is calculated by taking the min/avg/max score of the 8 references for each input sentence first, and then averaging across all scores.}
\vspace{-6pt}
\label{table:generation_model_comparison_after_bleu_filtering}
\end{table}

\subsection{Generation Models}
We consider two autoregressive language models, \textbf{GPT-2} \cite{Radford2019LanguageMA} and \textbf{GPT-3}\footnote{We use text-davinci-002, which is the most capable GPT-3 model.} \cite{NEURIPS2020_1457c0d6}, and two encoder-decoder language models, \textbf{BART} \cite{lewis-etal-2020-bart} and \textbf{T5} \cite{raffel2020exploring}.
For GPT-3, we try both zero-shot and few-shot (4 examples) setups using in-context learning without any fine-tuning.
For other models, we fine-tune seven configurations of them on \newdata.
Table \ref{table:generation_model_comparison_after_bleu_filtering} shows the test set results of each model and the diversity of human references measured by Self-BLEU.
Among all models, the few-shot setting of GPT-3 achieves the highest BERT-iBLEU score, and the zero-shot setting achieves the second-best number with only 1 point behind, which is not surprising given its size.
Compared to GPT-3 generations, human references are much more diverse with a decrease of 24.5 in Self-BLEU under the best case and 13.5 under the average case, indicating that there is still a big gap between large language models and humans.
For supervised small-scale models, T5\textsubscript{large} outperforms others with the best Self-BLEU and BERT-iBLEU scores.
Although BART\textsubscript{large} gets the highest BLEU score, our experiments in Appendix \ref{sec:further_generation_experiment} show BERT-iBLEU has the best correlation with human evaluation.
We thus use T5\textsubscript{large} in all the rest experiments.
For all models except GPT-3, we use beam search with beam size $=4$.
Please refer to Appendix \ref{sec:implementation_details} for details on the training setup and hyperparameter tuning. GPT-3 prompting and hyperparameter setup are provided in Appendix \ref{sec:gpt3}. Generation examples are displayed in Figure \ref{table:generation_examples_t5_gpt3} in Appendix \ref{appendix:examples}.



\setlength{\tabcolsep}{2pt}
\begin{table*}[t]
    \centering
    {\renewcommand{\arraystretch}{1.1}
    \resizebox{\textwidth}{!}{%
    \begin{tabular}{@{}lcccccc|cccc|cccc|cccc@{}}
        \toprule
        \multicolumn{3}{l}{\multirow{2}{*}{\diagbox[height=3.1em,width=8.5em]{\textit{Training set}}{\textit{Test set}}}}
          & \multicolumn{4}{c}{\textbf{\newtestdata}} & \multicolumn{4}{c}{\textbf{Quora}} & \multicolumn{4}{c}{\textbf{MSCOCO}} &
         \multicolumn{4}{c}{\textbf{ParaNMT}}
            \\ \cmidrule(lr){4-7} \cmidrule(lr){8-11} \cmidrule(lr){12-15} \cmidrule(lr){16-19}
        & & & BL & S-B $\downarrow$ & B-S & B-iB & BL & S-B $\downarrow$ & B-S & B-iB & BL & S-B $\downarrow$ & B-S & B-iB & BL & S-B $\downarrow$ & B-S & B-iB\\ 
       \midrule
            &
            \multicolumn{1}{l}{\textbf{\newdata}} & & \textbf{41.14} & 33.34 & \underline{85.86} & \textbf{77.79} & 26.28  & 46.98 & \underline{91.73} & \underline{67.31} & 19.69 & 56.59 & \textbf{92.86} & 66.44 & \underline{14.32} & 42.69 & \underline{86.10} & \underline{70.56} \\
           & \multicolumn{1}{l}{\textbf{Quora}} & & 32.13 & \underline{32.48} & 83.24 & \underline{76.07} & \textbf{28.72} & \underline{34.23} & 87.97 & \textbf{73.54} & 15.37 & 51.15 & 88.28 & 61.65 & 8.70 & \underline{28.73} & 79.79 & 67.67\\
           & \multicolumn{1}{l}{\textbf{MSCOCO}} & & 8.37 & \textbf{4.83} & 59.25 & 63.47 & 0.97 & \textbf{1.26} & 56.52 & 61.55 & \textbf{26.14} & \textbf{15.46} & 81.00 & \textbf{80.30} & 0.70 & \textbf{0.59} & 55.52 & 60.56 \\
           & \multicolumn{1}{l}{\textbf{ParaNMT}} & & \underline{38.69} & 47.74 & \textbf{90.98} & 75.66 & \underline{28.20} & 52.77 & \textbf{93.13} & 64.66 & \underline{19.75} & \underline{49.36} & \underline{92.59} & \underline{73.70} & \textbf{20.36} & 33.35 & \textbf{86.90} & \textbf{77.51} \\
        \bottomrule
    \end{tabular}
    }}
    \caption{Automatic evaluation of models fine-tuned on four datasets. Here, BL: BLEU, S-B: Self-BLEU, B-S: BERT-Score, B-iB: BERT-iBLEU. \textbf{Bold}: the best, \underline{Underline}: the second best.}
    \label{table:generation_dataset_comparison}
\end{table*}
\begin{figure}[t]

\begin{center}
\includegraphics[width=0.49\linewidth]{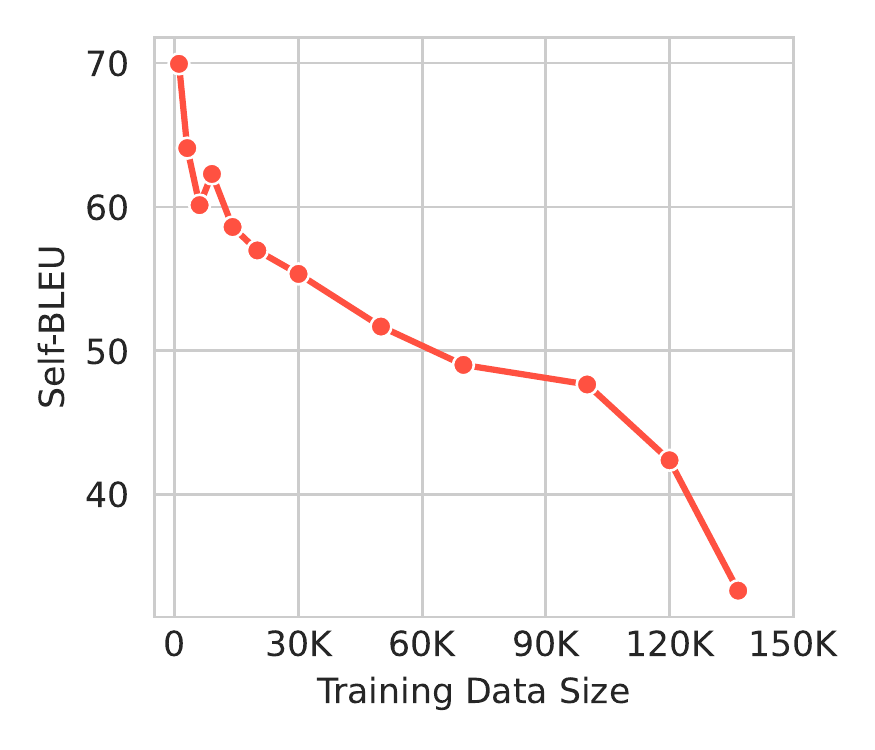}
\includegraphics[width=0.49\linewidth]{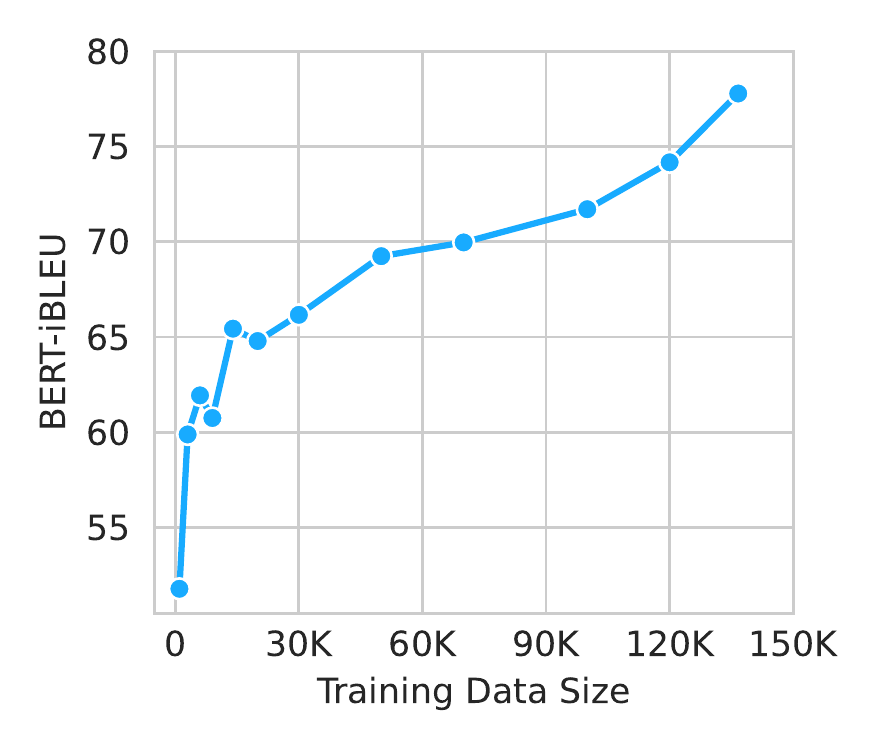}
\end{center}
\vspace{-10pt}
\caption{
Test set performance of model fine-tuned on varying amount of data in \newdata, in terms of Self-BLEU (lower is better) and BERT-iBLEU.
}
\vspace{-12pt}
\label{fig:datasize_vs_performance}
\end{figure}

\paragraph{Impact of Data Size.}
Figure \ref{fig:datasize_vs_performance} shows test set performance of T5\textsubscript{large} fine-tuned on different amount of data in \newdata~from 1K to 137K.
With more training data, the model generates more diverse and high-quality paraphrases as Self-BLEU decreases (improves) and BERT-iBLEU increases. 
This suggests that the paraphrase generation models will benefit from the continually growing size of  our \newdata~corpus.

\subsection{Cross-Dataset Generalization}
Building a paraphrase generation model that generalizes to new data is always an ambitious goal. To better understand the generalizability of each dataset, we fine-tune T5\textsubscript{large} on \newdata, Quora, MSCOCO, and ParaNMT separately and evaluate their performance across datasets.
For fair comparisons, we use the same architecture, T5\textsubscript{large}, in this experiment.
Appendix \ref{appendix:examples} displays examples generated by these models on each dataset.

Table \ref{table:generation_dataset_comparison} presents \textbf{automatic} evaluation of test set performance  across all four datasets.
As \newdata~and ParaNMT consist of sentences in different styles, models trained on them have better generalizability, achieving the best cross-domain performance.
On the contrary, since Quora and MSCOCO contain only questions or captions, models fine-tuned on them always generate question- or description-style sentences.
For example, given \textit{``we should take shots."}, model fine-tuned on Quora generates \textit{``Why do we take shots?"}.

\begin{table}[t]
\centering
\renewcommand{\arraystretch}{1}
\resizebox{0.95\linewidth}{!}{
\begin{tabular}{@{}lccc@{}}
\toprule
 \multirow{2}{*}{\textbf{Model}} & \multirow{2}{*}{\textbf{Fluency}} & \multicolumn{1}{p{2.4cm}}{\centering \textbf{Semantic Similarity}}  & \multirow{2}{*}{\textbf{Diversity}} \\
 \midrule
 \newdata & \textbf{4.98} & \textbf{4.67} & \textbf{3.59} \\
 ParaNMT & 4.95 & 4.64 & 3.40 \\

\bottomrule
\end{tabular}}
\captionof{table}{Human evaluation results on generations by model fine-tuned on \newdata~or ParaNMT.}
\vspace{-14pt}
\label{table:human_evaluation}
\end{table}

\begin{figure*}[ht]
\centering
\begin{subfigure}{.33\textwidth}
  \centering
  \includegraphics[width=\linewidth]{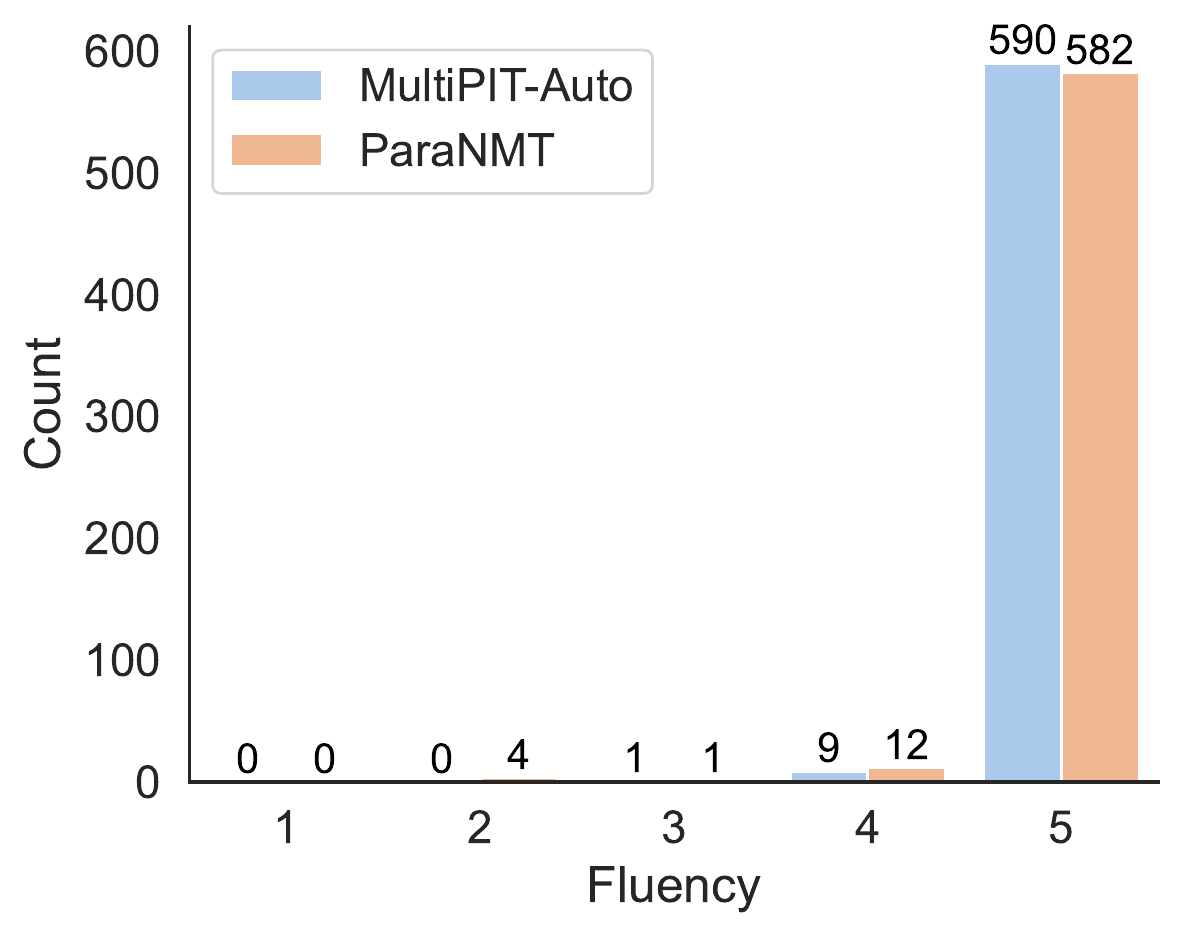}
\end{subfigure}%
\begin{subfigure}{.33\textwidth}
  \centering
  \includegraphics[width=\linewidth]{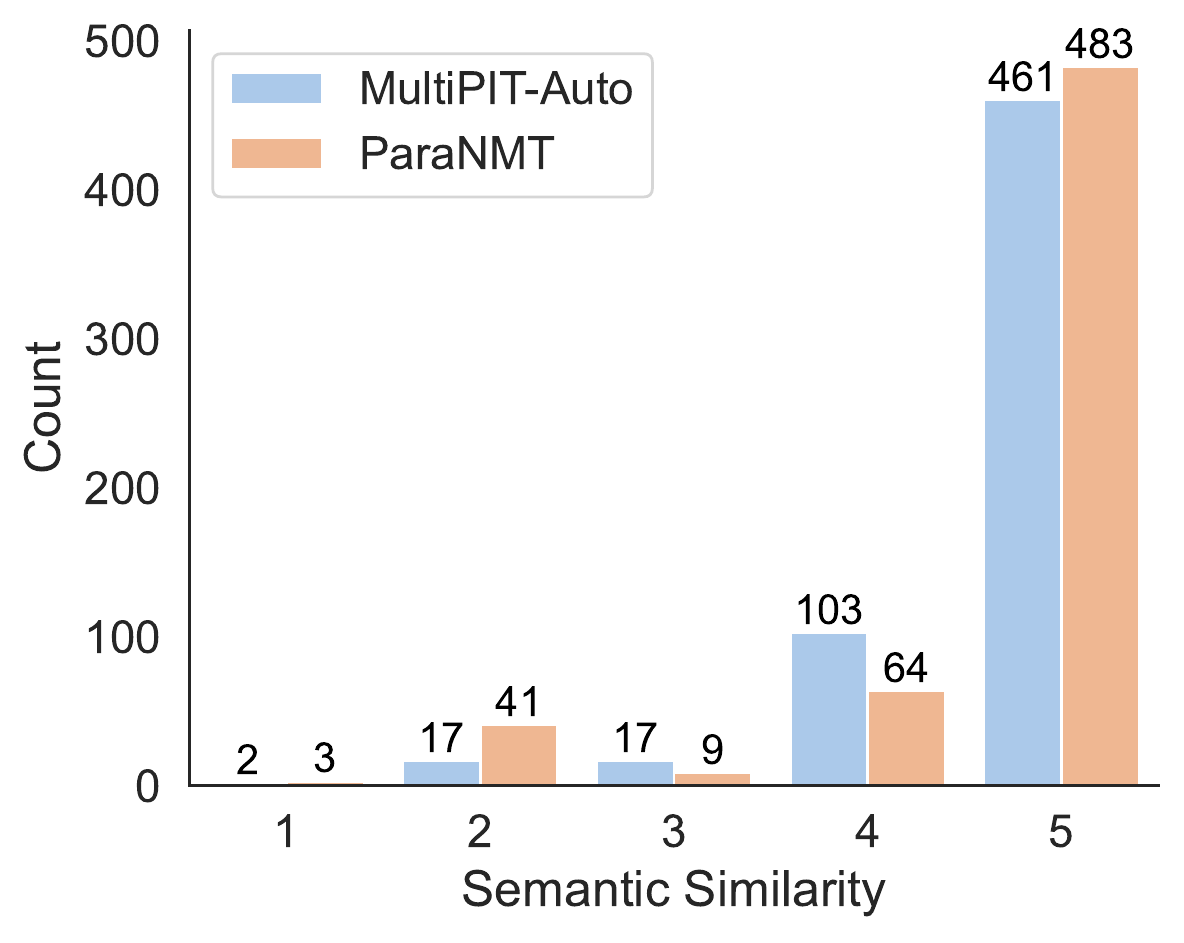}
\end{subfigure}
\begin{subfigure}{.33\textwidth}
  \centering
  \includegraphics[width=\linewidth]{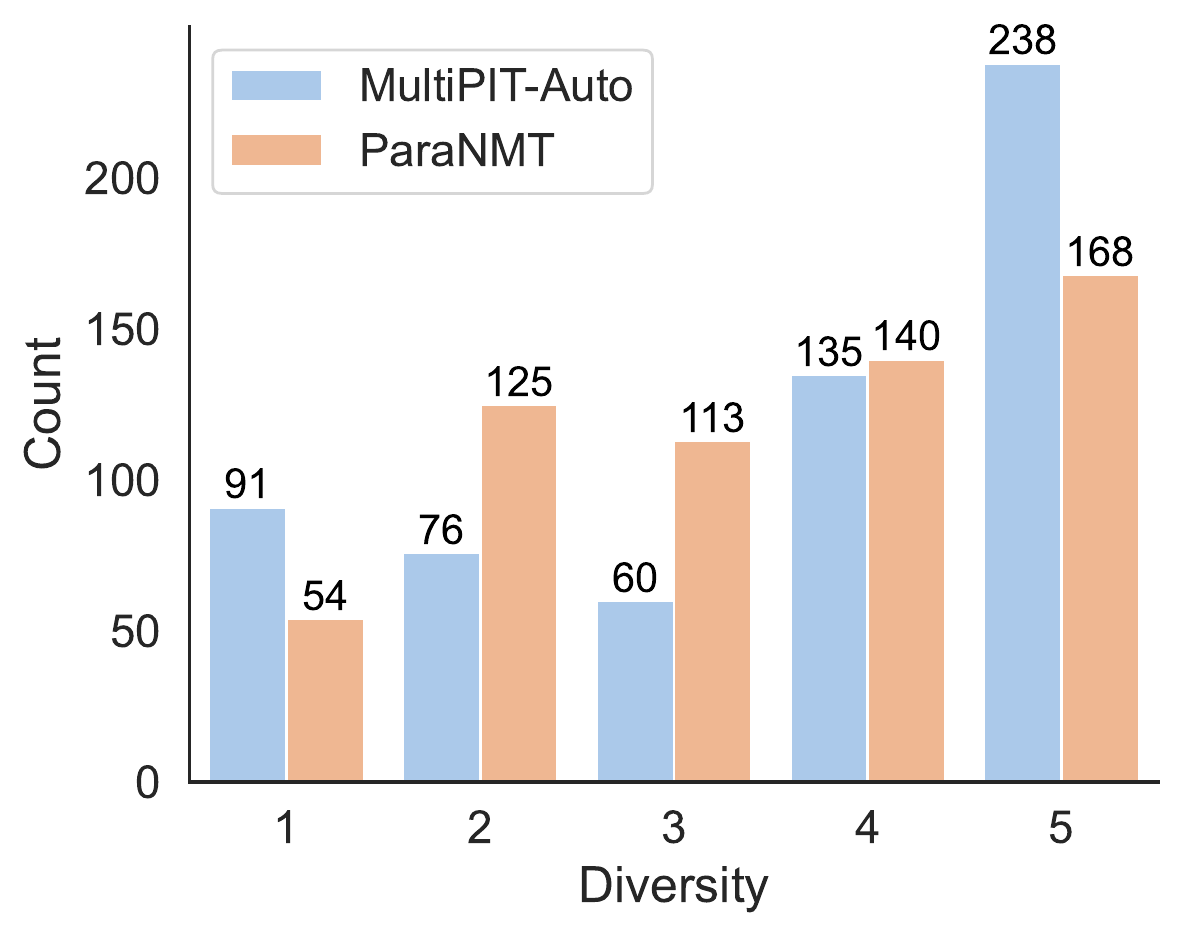}
\end{subfigure}
\vspace{-19pt}
\caption{
Human evaluation distributions on generations by model fine-tuned on \newdata~or ParaNMT.
}
\vspace{-8pt}
\label{fig:human_eval_label_distribution_by_model}
\end{figure*}
\setlength{\tabcolsep}{3pt}
\renewcommand{\arraystretch}{1.1} 
\begin{table*}[ht]
\centering
\small
\resizebox{\textwidth}{!}{%
\begin{tabular}{lp{0.25\textwidth}p{0.48\textwidth}cc}
\toprule
\textbf{Type} & \textbf{Definition} & \textbf{Generation Example} & \textbf{\newdataAbb} & \textbf{ParaNMT} \\ \midrule
\multicolumn{2}{c}{{\textit{\textbf{Good Paraphrase Type}}}} & \multicolumn{1}{r}{{Number of occurrences per generation:}} & \textbf{1.53} & 1.22 \\ \midrule
\nameAdd & Add new phrases while keeping the meaning of given sentence. & \textbf{Sent:} relax, take it easy. \textbf{Gen:} Relax, take a \paraAdd[deep breath, and enjoy the moment.] & \textbf{0.18} & 0.03 \\

\nameSynonym & Substitute a word with its synonym (another word). & \textbf{Sent:} \paraSynonym[Which] is the best GRE \paraSynonym[coaching] centre in Bangalore? \textbf{Gen:} \paraSynonym[what] is the best gre \paraSynonym[training] centre \ldots  & 0.39 & \textbf{0.54} \\

\namePos & Replace a phrase with synonym or expand a word to a phrase. & \textbf{Sent:} it \paraPos[looks goddamned foolish] to put an oyster on the clam. \textbf{Gen:} Putting an oyster on a clam \paraPos[is a fucking joke.]  & \textbf{0.28} & 0.16 \\

\nameStructure & Use different sentence structures to express the
same thing. & \textbf{Sent:} Two big plates \paraStructure[filled with some tasty looking] food. \textbf{Gen:} Two big plates of food, \paraStructure[and the food looks good.]
& \textbf{0.28} & 0.23 \\

\nameSimplification & Delete minor details or shorten phrases while maintaining the meaning of given sentence. & \textbf{Sent:} Daniel Farke sacked by Norwich after first win of Premier League season \paraSimplification[over Brentford.] \textbf{Gen:} Norwich sack Daniel Farke after first win of Premier League season. & \textbf{0.39} & 0.26  \\
\midrule
\multicolumn{2}{c}{{\textit{\textbf{Poor Paraphrase Type}}}} & \multicolumn{1}{r}{{Percentage in 200 generations:}} & 26\% & \textbf{44\%} \\ \midrule
\nameCopy & Copy the given sentence. & \textbf{Sent:} Did you have a good day today? \textbf{Gen:} Did you have a good day today? & \textbf{14.5\%} & 10\% \\

\nameSimple & Only have small changes such as changing article, tense, or prep. & \textbf{Sent:} FDA approves new test that can detect coronavirus in 45 minutes. \textbf{Gen:} \paraSimple[the] fda \paraSimple[has approved] \paraSimple[a] new test \ldots & 5.5\% & \textbf{18\%} \\

\nameHallu & Add new info that is not commonsense or world knowledge. & \textbf{Sent:} A dog at a table wearing a birthday hat. \textbf{Gen:} A dog wearing a birthday hat at a \paraHallu[dinner party.] & \textbf{2.5\%} & 0\% \\

\nameMiss & Miss important info in the given sentence. & \textbf{Sent:} \paraMiss[Very sad though] that the amazing AJ and Kai will be missing the final. \textbf{Gen:} AJ and Kai will not be in the final. & \textbf{1.5\%} & 1\% \\

\nameMisinterpret & Misinterpret or contradict meaning of the given sentence. & \textbf{Sent:} Why are most first basemen left handed? \textbf{Gen:} why do most of the first basemen \paraMisinterpret[have left hands?] & 2\% & \textbf{14\%} \\
\nameGrammar & Contain grammar error. & \textbf{Gen:} what \paraGrammar[is] the best earphones for rs 3000? & 0\% & \textbf{1\%} \\
\bottomrule
\end{tabular}%
}
\caption{Paraphrase types with examples and statistics observed in the generations by models fine-tuned on \newdata~(\newdataAbb)~or ParaNMT. Statistics are based on manual inspection of generations by each model on 200 sampled sentences. The shown generation example for each type is by model with the higher value (\textbf{bold}).}
\vspace{-10pt}
\label{tab:manual_inspection}
\end{table*}

We conduct a \textbf{human evaluation}  to further compare \newdata~and ParaNMT datasets, by evaluating 200 randomly sampled generations from  the model trained on each corpus.\footnote{The input is 4 $\times$ 50 sentences from each test set.}
As shown in Table \ref{table:human_evaluation}, \newdata's generations receive the highest scores in all three dimensions: \textit{fluency}, \textit{semantic similarity} and \textit{diversity}.
Each generation is rated by three annotators on a 5-point Likert-scale per aspect, with 5 being the best.
We also show the distribution of human evaluation results on each dimension in Figure \ref{fig:human_eval_label_distribution_by_model} for a deeper comparison.
Specifically, \newdata~model generates fewer really poor paraphrases (semantic similarity $<$3) and much more diverse paraphrases (diversity $>$3).
We include our evaluation template in Appendix \ref{sec:human_evaluation}.
We measure inter-annotator agreement using ordinal Krippendorff's alpha \cite{Krippendorff2011ComputingKA}, which yields 0.31 for fluency,\footnote{Since over 95\% ratings of fluency fall into the same point (see Figure \ref{fig:human_eval_label_distribution} in the Appendix), Krippendorff's alpha will stay low no matter how often the raters agree.} 0.56 for semantic similarity, and 0.81 for diversity. All values are considered fair to good \cite{krippendorff2004reliability}.

Additionally, we perform a \textbf{manual inspection} and observe that model fine-tuned on \newdata~generates more diverse kinds of good paraphrases and much fewer poor paraphrases than the one trained on ParaNMT.
We define five good paraphrase types and six poor paraphrase types.
The definitions and results are shown in Table \ref{tab:manual_inspection}.


\section{Other Related Work}
Besides the several frequently used paraphrase datasets we mentioned above, here are a few other paraphrase corpora.
The MSR Paraphrase corpus \cite{dolan-brockett-2005-automatically} contains 5,801 sentences pairs from news articles, but it has a deficiency that skewing toward over-identification \cite{das-smith-2009-paraphrase} and having high lexical overlap \cite{Rus2014OnPI}.
%
PPDB \cite{Ganitkevitch2013PPDBTP} contains over 220 million phrase and lexical paraphrases without any sentence paraphrases.
WikiAnswer \cite{Fader2013ParaphraseDrivenLF} consists of 18 million word-aligned question pairs. 
However, same as Quora, WikiAnswer is restricted to only questions.
In addition, the Semantic Textual Similarity (STS) shared task \citet{cer-etal-2017-semeval} measures the degree to which two sentences are semantically similar to each other. Since it doesn't make a binary judgment for paraphrase relationships, it is not frequently used in paraphrase research.
Recently, \citet{dong2021parasci} presents ParaSci, a large paraphrase dataset in the scientific field, and \citet{kim-etal-2021-bisect} proposes BiSECT, a large split and rephrase corpus constructed using machine translation.
Our work focuses on creating a large paraphrase corpus that contains more diverse and natural human-authored texts and investigating different paraphrase criteria.

 

\section{Conclusion}
In this paper, we present the Multi-Topic Paraphrase in Twitter (\datasetname) corpus. Our work surpasses prior Twitter-based paraphrase corpora in topic diversity as well as the  quality and quantity of annotation.  Experimental results demonstrate the necessity of defining paraphrases based on downstream tasks. 
Our paraphrase generation evaluation shows that models trained on our corpus have better generation quality and generalizability   compared to models fine-tuned on existing widely-used paraphrase datasets.
We believe that \datasetname~will facilitate further research in both paraphrase identification and paraphrase generation.

\section*{Limitations}
While our study shows \newdata~improves paraphrase generation quality and diversity, we observe model sometimes generates Twitter-specific artifacts (i.e. \textit{``@JoeBiden"}).
Future work could investigate techniques to mine paraphrases from other social media platforms such as Reddit.
Another limitation is that our dataset is only in English, future work could extend this to multilingual as Twitter is used by users from different countries that speak different languages.

\section*{Acknowledgments}


We thank Yang Chen as well as three anonymous reviewers for their helpful feedback on this work. We also thank Andrew Duffy, Elizabeth Liu, Ian Ligon, Rachel Choi, Jonathan Zhou, Chase Perry, Panya Bhinder for their help on annotations and human evaluation. This research is supported in part by the NSF awards IIS-2144493 and IIS-2112633, ODNI and IARPA via the BETTER program (contract 19051600004), and Figure
Eight AI for Everyone Award. The views and conclusions contained herein are those of the authors and should not be interpreted as necessarily representing the official policies, either expressed or implied, of NSF, ODNI, IARPA, or the U.S. Government. The U.S. Government is authorized to reproduce and distribute reprints for governmental purposes notwithstanding any copyright annotation therein.
\bibliographystyle{acl_natbib}
\bibliography{custom}

\appendix

\clearpage
\section{Annotation Interface}
\subsection{Crowdsourcing}
\label{appendix:crowd-annotation}
Figure \ref{fig:crowd_annotation_interface} and Figure \ref{fig:crowd_annotation_question} display screenshots of the instruction and an example question of our crowdsourcing annotation for \crowddata.

\subsection{Expert}
\label{appendix:expert-annotation}
Figure \ref{fig:expert_annotation_interface} displays a screenshot of the instruction of our expert annotation for \expertdata.


\section{Data Pre-processing}
\label{appendix:preprocessing}

Both PIT-2015 \cite{xu2015semeval} and Twitter URL \cite{lan-etal-2017-continuously} datasets share similar pre-processing steps that introduced tokenization and sentence splitting errors. Moreover, PIT-2015 contains some spam patterns, such as ``Follow Me PLEASE''.  We improved the quality of our dataset by fixing the pre-processing methods and removing spam patterns. More importantly, we split tweets into sentences to get cleaner paraphrases (see Table \ref{tab:raw_tweets} for an example), without added noises from extra sentences in the tweet. We improve the sentence splitting script by \citet{xu2015semeval} and tokenization script by \citet{o2010tweetmotif} used in  prior work with a number of errors fixed: (1) Emojis and most symbols are cleaned while punctuation are kept; (2) Extremely short sentences ($<5$ tokens) are filtered out while remaining sentences are deduplicated by comparing lowercased strings  w/o any punctuation.

\begin{table}[h]
\small
    \centering
    \resizebox{0.98\columnwidth}{!}{%
    \begin{tabular}{ll}
        \toprule
        \multicolumn{2}{l}{\enskip\textbf{Raw Tweets w/o Sent. Splitting}}  \\ \midrule
        $\circ$ & \textit{Horrible Crash on the Aurora Bridge in Seattle.}  \\
        $\circ$ & \textit{The crash on the Aurora Bridge in Seattle looks horrible}.\\
        &  \textit{That was the bridge I took to work everyday. Yikes.} \\ \toprule
    \end{tabular}}
    \vspace{-6pt}
    \caption{An example pair of raw tweets from our corpus. Annotating at tweet-level will include mismatched content and ambiguity. Cleaner paraphrase annotations can be acquired after sentence splitting.}
    \vspace{-12pt}
    \label{tab:raw_tweets}
\end{table}

\section{Implementation Details}
\label{sec:implementation_details}
We use HuggingFace Transformers \cite{wolf-etal-2020-transformers} version of all pre-trained models.
We use Python 3.8, PyTorch 1.9.0, and Transformers 4.12.0.
For all experiments, we use 4 $\times$ 48GB NVIDIA A40 GPUs.

\paragraph{Paraphrase Identification.}
Hyperparameters for fine-tuning models in paraphrase identification experiments are given in Table \ref{table:hyperparameter_identification}.

\begin{table}[h]
\centering
\renewcommand{\arraystretch}{1}
\resizebox{0.9\linewidth}{!}{
\begin{tabular}{@{}cc@{}}
\toprule
 \textbf{Hyperparameter} & \textbf{Assignment}\\
 \midrule
 Max epochs & 5  \\
 Eval steps & 500 \\
 Effective batch size & 32 \\
 Learning rate optimizer & AdamW \\
 Adam epsilon & 1e-8 \\
 Weight decay & 0.01 \\
 Learning rate & \{1e-5, 2e-5, 3e-5, 5e-5\} \\
 Learning rate decay & Linear \\
 Warmup ratio & 0.06 \\
\bottomrule
\end{tabular}}
\captionof{table}{Hyperparameters for paraphrase identification. We choose learning rate range based on \citet{Liu2019RoBERTaAR}}
\vspace{-6pt}
\label{table:hyperparameter_identification}
\end{table}

For T5 model, we consider learning rates $\in$ \{1e-4, 3e-4, 1e-5, 3e-5\}. For DeBERTaV3 model, we consider learning rates $\in$ \{1e-5, 3e-5, 5e-6, 8e-6\} following \citet{He2021DeBERTaV3ID}.
We fine-tune for 5 epochs and eval every 500 steps (every epoch if total training steps is less than 1500) on the dev set. 
The only hyperparameter we tune is the learning rate and use $F_1$ on the dev set for model selection.


For Infersent and ESIM models, we use their original implementation initialized with GloVe embedding \cite{pennington-etal-2014-glove}, and also only tune the learning rate based on the dev set. 

\paragraph{Paraphrase Generation.}
Hyperparameters for fine-tuning models in paraphrase generation experiments are given in Table \ref{table:hyperparameter_generation}.

\begin{table}[h]
\centering
\renewcommand{\arraystretch}{1}
\resizebox{0.9\linewidth}{!}{
\begin{tabular}{@{}cc@{}}
\toprule
 \textbf{Hyperparameter} & \textbf{Assignment}\\
 \midrule
 Max epochs & 5  \\
 Eval steps & 500 \\
 Effective batch size & 128 \\
 Learning rate optimizer & AdamW \\
 Adam epsilon & 1e-8 \\
 Weight decay & 0.01 \\
 Learning rate & \{1e-4, 3e-4, 1e-5, 3e-5\} \\
 Learning rate decay & Linear \\
 Warmup ratio & 0.06 \\
\bottomrule
\end{tabular}}
\captionof{table}{Hyperparameters for paraphrase generation.}
\vspace{-6pt}
\label{table:hyperparameter_generation}
\end{table}

We use perplexity on the dev set for model selection.

As ParaNMT contains only lowercase letters, we lowercase the input and references for generation and evaluation of the model fine-tuned on ParaNMT and lowercase the other models' generations while evaluating on ParaNMT.
\section{GPT-3 Setup}
\label{sec:gpt3}
\subsection{Hyperparameters}
We use the text-davinci-002 GPT-3 model for paraphrase generation. To generate paraphrase, we use the following hyperparameters: temperature=1, max tokens=100, top-p=0.9, best of=1, frequency penalty=0.5, presence penalty=0.5, based on \citet{chakrabarty2022flute}.

\subsection{Prompts}
\paragraph{Zero-shot setting:}
\textit{Your task is to generate a diverse paraphrase for a given sentence. \\
\\
Sentence: \{sentence\} \\
Paraphrase:}

\paragraph{Few-shot setting:}
\textit{
You will be presented with examples of some input sentences and their paraphrases. Your task is to generate a diverse paraphrase for a given sentence.\\
\\
Sentence: Mike Bloomberg is sending \$18 million from his defunct presidential campaign to the DNC .\\
Paraphrase: Mike Bloomberg is transferring \$18M from his campaign to DNC , stretching campaign finance law .\\
\\
Sentence: Google Assistant on Android can read web pages to you\\
Paraphrase: Google Assist lets your Android devices read entire web pages aloud\\
\\
Sentence: Charlie Patino scored a goal on his debut !\\
Paraphrase: Charlie Patino's debut and he capped it off with a goal .\\
\\
Sentence: khem birch is the difference maker for the raptors this game\\
Paraphrase: Khem Birch may be the MVP tonight for the Raptors .\\
\\
Sentence: \{sentence\}\\
Paraphrase:
}

\section{Generation Dataset Statistics}
\label{sec:details:statistics}
Table \ref{table:paraphrase_generation_data_stats} presents the detailed statistics of \newdata, Quora, MSCOCO and ParaNMT.

\begin{table}[ht]
\centering
\renewcommand{\arraystretch}{1.1}
\resizebox{\linewidth}{!}{
\begin{tabular}{lcccc}
\toprule
 & \textbf{\newdataAbb} & \textbf{Quora} & \textbf{MSCOCO} & \textbf{ParaNMT} \\
 \midrule
    Genre & Twitter & Question & Description & Novels, Laws \\
    Sentence Length & 11.34 & 9.66  & 10.49  & 11.33\\
 Sentence BLEU & 24.48 & 26.37 & 9.30 & 24.85\\ \midrule
 \multicolumn{5}{l}{\textit{Train/dev/test split}} \\
 \#Train w/o BF & 290,395 & 134,378 & 331,330 & 50M \\
 \#Train & 136,645 & 47,393 & 275,583 & 443,512 \\
 \#Dev &  215  &   5,255   & 20,186 & 499\\
\#Test & 200 & 5,255 & 20,187 & 781 \\
\#Test Refs & 8 & 1.34 & 4 & 1 \\
\bottomrule
\end{tabular}}
\caption{Statistics of datasets for paraphrase generation. We calculate sentence length based on the number of tokens per unique sentence. As ParaNMT is too large, we sample 500K for the calculation of sentence length and BLEU. W/o BF denotes without BLEU filtering.}
\label{table:paraphrase_generation_data_stats}
\end{table}
\section{Further Paraphrase Generation Experiments}
\label{sec:further_generation_experiment}

\begin{table}[h]
\renewcommand{\arraystretch}{1}
\small
\centering
\resizebox{\linewidth}{!}{
\begin{tabular}{lllll}
\toprule
\textbf{Metric} & \textbf{Fluency} & \textbf{Semantic} & \textbf{Diversity} & \textbf{Overall} \\
    & & \textbf{Similarity} & & \\
\midrule
   BLEU & 0.212 & 0.209 & -0.233 & -0.091 \\
   Self-BLEU $\downarrow$ & 0.068 & 0.412$^{***}$ & -0.655$^{***}$ & -0.452$^{***}$ \\
   BERT-Score & 0.062 & 0.523$^{***}$ & -0.722$^{***}$ & -0.507$^{***}$ \\
   BERT-iBLEU & -0.166 & -0.089 & 0.370$^{**}$ & 0.381$^{***}$ \\
    \bottomrule
\end{tabular}
}
\caption{Spearman correlations with human evaluation on 100 generations on \newtestdata~(50 by model trained on \newdata~and 50 by model trained on ParaNMT). Here, $^{***}$: p < 0.0001, $^{**}$: p < 0.001, $^*$: p < 0.01. Overall is the summation score of all three aspects.}
\label{table:spearman_correlation_multipit}
\end{table}
\begin{table}[h]
\renewcommand{\arraystretch}{1}
\small
\centering
\resizebox{\linewidth}{!}{
\begin{tabular}{lllll}
\toprule
\textbf{Metric} & \textbf{Fluency} & \textbf{Semantic} & \textbf{Diversity} & \textbf{Overall} \\
    & & \textbf{Similarity} & & \\
\midrule
   Self-BLEU $\downarrow$ & 0.043 & 0.319$^{***}$ & -0.638$^{***}$ & -0.491$^{***}$ \\
   BERT-Score & 0.070 & 0.436$^{***}$ & -0.744$^{***}$ & -0.561$^{***}$ \\
   BERT-iBLEU & -0.036 & -0.096 & 0.346$^{***}$ & 0.339$^{***}$ \\
    \bottomrule
\end{tabular}
}
\caption{Spearman correlations with human evaluation on all 400 generations. Here, $^{***}$: p < 0.0001, $^{**}$: p < 0.001, $^*$: p < 0.01.}
\label{table:spearman_correlation_whole}
\end{table}
\begin{figure}[ht]

\begin{center}
\includegraphics[width=0.99\linewidth]{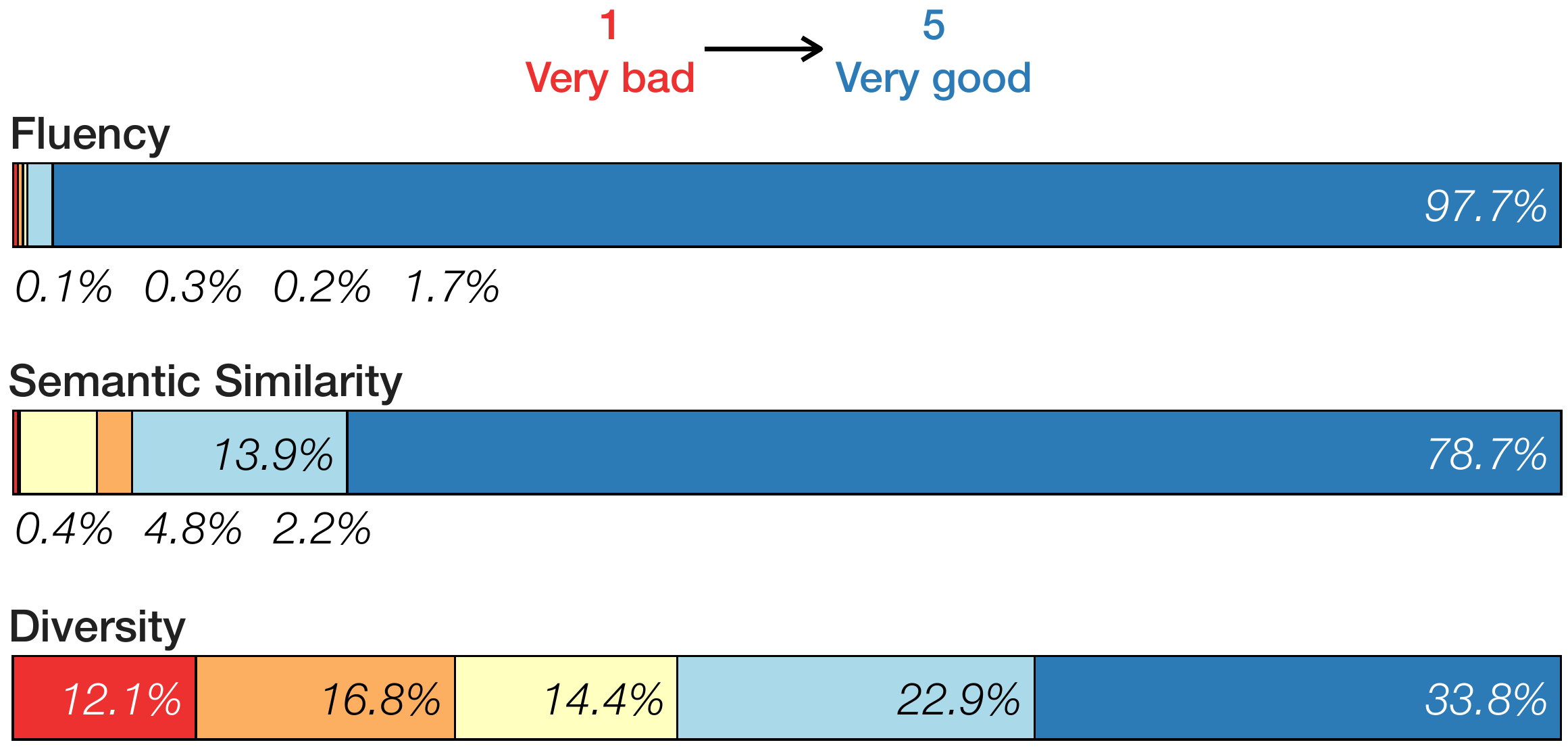}
\end{center}
\vspace{-5pt}
\caption{
Label distribution of 1200 ratings on 400 generations by models fine-tuned on \newdata~and ParaNMT.
}
\label{fig:human_eval_label_distribution}
\end{figure}

\paragraph{Correlation Analysis.}
With human evaluation, we calculate Spearman correlation to evaluate automatic metric quality. Since the four test sets have different numbers of references and \newtestdata~has the most number of references, to evaluate BLEU, we examine 100 generations on \newtestdata~(50 by T5\textsubscript{large} fine-tuned on \newdata~and 50 by T5\textsubscript{large} fine-tuned on ParaNMT).
Results are shown in Table \ref{table:spearman_correlation_multipit}.
BLEU gets a weak correlation around $|0.2|$ with all aspects and $\sim$0.1 with the overall score.
Table \ref{table:spearman_correlation_whole} presents Spearman correlations for Self-BLEU, BERT-Score and BERT-iBLEU on all 400 generations. BERT-iBLEU outperforms the other two metrics. 
Because Self-BLEU measures diversity and BERT-Score measures semantic similarity, both metrics get the best correlation with human evaluation on the corresponding aspect but the worst correlation on the other one. Notably, Self-BLEU gets the highest correlation with the overall measurement, but the reason behind it is more differentiation in diversity ratings compared to semantic similarity, as shown in Figure \ref{fig:human_eval_label_distribution}. This makes diversity the biggest role in the overall score.

\begin{figure}[h]

\begin{center}

\includegraphics[width=0.49\linewidth]{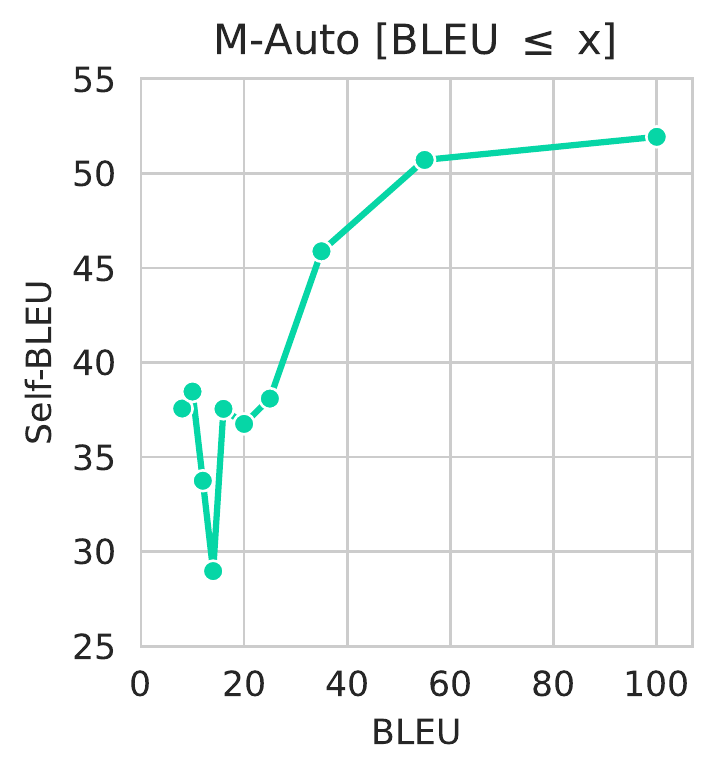}
\includegraphics[width=0.49\linewidth]{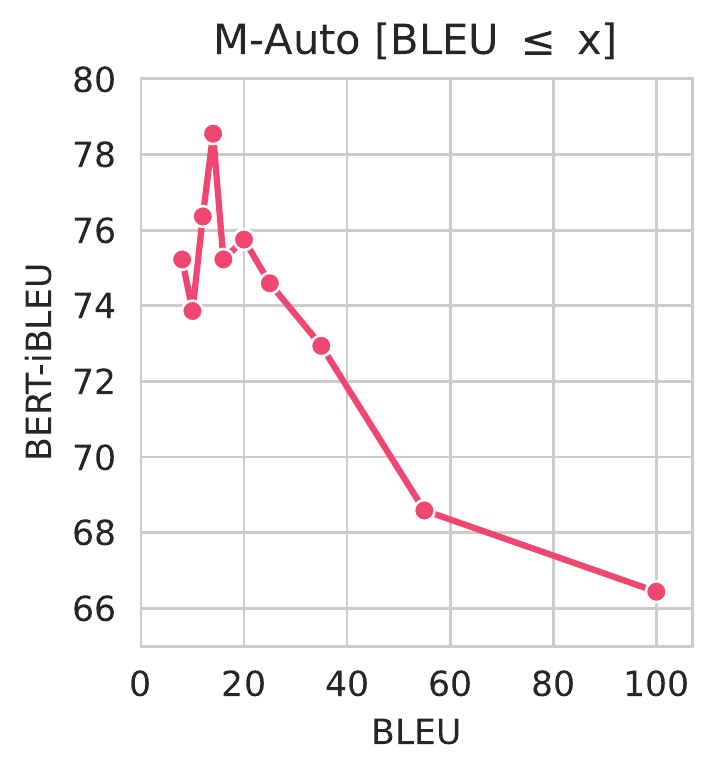}

\end{center}

\caption{
\newdata~dev set performance on various BLEU filtering thresholds.
}
\label{fig:best_setup}
\end{figure}
\begin{table}[h]
\renewcommand{\arraystretch}{1.1}
\small
\centering
\resizebox{\linewidth}{!}{
\begin{tabular}{lccccc}

\toprule
Training Data & LR & BL & S-B $\downarrow$ & B-S & B-iB \\
    \midrule
     \newdataAbb~w/o BF & 1e-4 & 45.85 & 65.03 & 91.86 & 59.87 \\
      \newdataAbb & 1e-4 & 41.14 & 33.34 & 85.86 & \textbf{77.79}  \\ \midrule
      Quora w/o BF & 3e-4 & 36.63 & 54.54 & 91.27 & 61.58 \\
      Quora & 1e-4 & 28.72 & 34.23 & 87.97 & \textbf{73.54} \\ \midrule
        MSCOCO w/o BF & 3e-4 &  28.15 & 23.39 & 82.99 & 78.89\\
      MSCOCO & 1e-4 & 26.14 & 15.46 & 81.00 & \textbf{80.30} \\ \midrule
        ParaNMT w/o BF & 1e-5 & 19.43 & 37.59 & 88.07 & 76.28\\
      ParaNMT & 3e-5 & 20.36 & 33.35 & 86.90 & \textbf{77.51} \\
    \bottomrule
\end{tabular}
}
\caption{In-domain test set results of fine-tuning model on data with or without BLEU filtering. w/o BF denotes without BLEU filtering.}
\label{table:bleu_filtering_datasets}
\end{table}

\paragraph{BLEU Filtering.}
We evaluate different BLEU thresholds on the dev set of \newdata~as shown in Figure \ref{fig:best_setup}.
The model achieves the best performance at the threshold of 14, which is used across our experiments.

Next, we compare model performance on all four datasets with and without BLEU filtering.
Results are presented in Table \ref{table:bleu_filtering_datasets}.
Applying BLEU filtering improves model performance with higher BERT-iBLEU on all datasets.

\paragraph{Impact of Definition.}
We investigate how different paraphrase definitions affect generation performance. As shown in Table \ref{table:generation_variations}, model fine-tuned on \newdata~outperforms fine-tuning on the loosely defined data such as \crowddata.

\begin{table}[ht]
\renewcommand{\arraystretch}{1.1}
\small
\centering
\resizebox{\linewidth}{!}{
\begin{tabular}{llcccc}

\toprule
Data & Size & BL & S-B $\downarrow$ & B-S & B-iB \\
    \midrule
    \crowddata & 26,091 & 36.15 & \textbf{32.09} & 85.53 & 74.19 \\
    \newdataAbb\textsubscript{\textsc{-crowd}} & 326,517 & \textbf{45.55} & 37.90 & 85.80 & 74.12 \\
    \newdataAbb & 136,645 & 41.14 & 33.34 & \textbf{85.86} & \textbf{77.79} \\
    
    \bottomrule
\end{tabular}
}
\caption{Test set results of models fine-tuned on data constructed with different paraphrase definitions. \crowddata~contains its paraphrase pairs. \newdataAbb\textsubscript{\textsc{-crowd}} is the automatically identified paraphrase pairs by the identifier fine-tuned on \crowddata.}
\label{table:generation_variations}
\end{table}

\section{Examples}
\label{appendix:examples}
\paragraph{Generation Examples.}
Table \ref{table:generation_examples_t5_gpt3} presents generation examples by GPT-3 and fine-tuned T5\textsubscript{large} on \newtestdata.

Table \ref{table:generation_examples} presents generation examples by T5\textsubscript{large} fine-tuned on \newdata, Quora, MSCOCO, and ParaNMT.

\paragraph{Multi-Reference Examples.}
Table \ref{table:multi_ref_examples} displays three examples from the \newtestdata~test set.

\begin{table*}[t]
\centering
\renewcommand{\arraystretch}{1.2}
\resizebox{\linewidth}{!}{
\begin{tabular}{l}
\toprule
																								
\textbf{Source sentence:} Today is the winter solstice, marking the beginning of winter in the Northern Hemisphere. \\
\textbf{Generated sentences: }\\ 
\textit{T5\textsubscript{large}} $\rightarrow$ Today marks the beginning of winter in the Northern Hemisphere. \\
\textit{GPT-3\textsubscript{zero-shot}} $\rightarrow$ Today is the first day of winter in the Northern Hemisphere. \\
\textit{GPT-3\textsubscript{few-shot}} $\rightarrow$ Today is the winter solstice, which means winter has arrived in the Northern Hemisphere. \\
\midrule
\textbf{Source sentence:} Elizabeth Warren and Cory Booker test positive for COVID in breakthrough cases \\
\textbf{Generated sentences: }\\ 
\textit{T5\textsubscript{large}} $\rightarrow$ Cory Booker and Elizabeth Warren have tested positive for COVID-19. \\
\textit{GPT-3\textsubscript{zero-shot}} $\rightarrow$ In a breakthrough, Elizabeth Warren and Cory Booker test positive for COVID. \\
\textit{GPT-3\textsubscript{few-shot}} $\rightarrow$ Elizabeth Warren and Cory Booker have both contracted COVID-19 in what are being called breakthrough cases. \\
\midrule
\textbf{Source sentence:} that malcolm subban save is f*cking incredible. \\
\textbf{Generated sentences: }\\ 
\textit{T5\textsubscript{large}} $\rightarrow$ Malcolm Subban with the save of the year. \\
\textit{GPT-3\textsubscript{zero-shot}} $\rightarrow$ That save by Malcolm Subban was insane! \\
\textit{GPT-3\textsubscript{few-shot}} $\rightarrow$ That Malcolm Subban save was absolutely incredible. \\
\midrule
\textbf{Source sentence:} Former Oklahoma quarterback Spencer Rattler is transferring to University of South Carolina. \\
\textbf{Generated sentences: }\\ 
\textit{T5\textsubscript{large}} $\rightarrow$ Former Oklahoma QB Spencer Rattler is transferring to the University of South Carolina. \\
\textit{GPT-3\textsubscript{zero-shot}} $\rightarrow$ Former Oklahoma quarterback Spencer Rattler is transferring to the University of South Carolina. \\
\textit{GPT-3\textsubscript{few-shot}} $\rightarrow$ former OU quarterback Spencer Rattler is transferring to the University of South Carolina. \\
\midrule
\textbf{Source sentence:} The Vancouver Canucks have fired head coach Travis Green and hired Bruce Boudreau as a replacement. \\
\textbf{Generated sentences: }\\ 
\textit{T5\textsubscript{large}} $\rightarrow$ Canucks fire Travis Green, hire Bruce Boudreau as new head coach \\
\textit{GPT-3\textsubscript{zero-shot}} $\rightarrow$ The Vancouver Canucks have replaced head coach Travis Green with Bruce Boudreau. \\
\textit{GPT-3\textsubscript{few-shot}} $\rightarrow$ The Vancouver Canucks have let go of head coach Travis Green and replaced him with Bruce Boudreau. \\
\midrule
\textbf{Source sentence:} What an incredible tribute to Eddie Kidd on Top Gear. \\
\textbf{Generated sentences: }\\ 
\textit{T5\textsubscript{large}} $\rightarrow$ What a great tribute to Eddie Kidd on Top Gear. \\
\textit{GPT-3\textsubscript{zero-shot}} $\rightarrow$ Top Gear did an amazing job honoring Eddie Kidd. \\
\textit{GPT-3\textsubscript{few-shot}} $\rightarrow$ What a fitting tribute to Eddie Kidd on Top Gear. \\
\midrule
\textbf{Source sentence:} How to build immunity for Secondhand Stress \\
\textbf{Generated sentences: }\\ 
\textit{T5\textsubscript{large}} $\rightarrow$ How can you build your immunity against secondhand stress? \\
\textit{GPT-3\textsubscript{zero-shot}} $\rightarrow$ There are many ways to build immunity against secondhand stress. Some methods include yoga, meditation, and aromatherapy. \\
\textit{GPT-3\textsubscript{few-shot}} $\rightarrow$ How to become immune to Secondhand Stress \\
\midrule
\textbf{Source sentence:} Twitter is trialling its own 24-hour Stories feature called Fleets \\
\textbf{Generated sentences: }\\ 
\textit{T5\textsubscript{large}} $\rightarrow$ @Twitter is testing a new 24-hour stories feature called Fleets. \\
\textit{GPT-3\textsubscript{zero-shot}} $\rightarrow$ Twitter is testing out a new 24-hour Stories feature called Fleets. \\
\textit{GPT-3\textsubscript{few-shot}} $\rightarrow$ Twitter is testing a new 24-hour Stories feature called Fleets. \\
\bottomrule
\end{tabular}}
\caption{Paraphrase generation examples by GPT-3 and fine-tuned T5\textsubscript{large} on \newtestdata.}
\label{table:generation_examples_t5_gpt3}
\end{table*}

\begin{table*}[t]
\centering
\renewcommand{\arraystretch}{1.2}
\resizebox{\linewidth}{!}{
\begin{tabular}{l}
\toprule
\textbf{Source origin:} \newtestdata \\
\textbf{Source sentence:} In honor of Veterans Day, we salute all of the men and women who have served in America's armed forces. \\
\textbf{Generated sentences: }\\ 
\textit{T5${_{\newdata}}$} $\rightarrow$ Today is Veterans Day, a day to remember those who have served. \\
\textit{T5${_{Quora}}$} $\rightarrow$ On this Veterans Day, we salute all the men and women who have served our country. \\
\textit{T5${_{MSCOCO}}$} $\rightarrow$ A group of men and women standing next to each other. \\
\textit{T5${_{ParaNMT}}$} $\rightarrow$ in honor of veterans'day, we salute all the men and women who served in america's armed forces. \\
\midrule
\textbf{Source origin:} Quora\\
\textbf{Source sentence:} What is the scope for a food technologist in India? \\
\textbf{Generated sentences: }\\ 
\textit{T5${_{\newdata}}$} $\rightarrow$ What is the job outlook for a food technologist in India? \\
\textit{T5${_{Quora}}$} $\rightarrow$ What is the scope of food technology in India? \\
\textit{T5${_{MSCOCO}}$} $\rightarrow$ a man in a white shirt is working in a kitchen \\
\textit{T5${_{ParaNMT}}$} $\rightarrow$ what is the role of a food technologist in india? \\
\midrule
\textbf{Source origin:} MSCOCO\\
\textbf{Source sentence:} A bear is taking a walk through the forest. \\
\textbf{Generated sentences: }\\ 
\textit{T5${_{\newdata}}$} $\rightarrow$ A bear is taking a walk in the woods. \\
\textit{T5${_{Quora}}$} $\rightarrow$ What is a bear doing in the forest? \\
\textit{T5${_{MSCOCO}}$} $\rightarrow$ A brown bear walking through a wooded area. \\
\textit{T5${_{ParaNMT}}$} $\rightarrow$ the bear is walking through the forest. \\
\midrule
\textbf{Source origin:} ParaNMT\\
\textbf{Source sentence:} someone speaks in english over there. \\
\textbf{Generated sentences: }\\ 
\textit{T5${_{\newdata}}$} $\rightarrow$ Someone has to speak english in this room. \\
\textit{T5${_{Quora}}$} $\rightarrow$ What is the best way to learn English from someone who speaks fluently? \\
\textit{T5${_{MSCOCO}}$} $\rightarrow$ A man standing in front of a wall with a sign on it. \\
\textit{T5${_{ParaNMT}}$} $\rightarrow$ someone's speaking english. \\
\bottomrule
\end{tabular}}
\caption{Paraphrase generation examples by T5\textsubscript{large} fine-tuned on \newdata, Quora, MSCOCO, and ParaNMT on each test set.}
\label{table:generation_examples}
\end{table*}

\begin{table*}[t]
\centering
\renewcommand{\arraystretch}{1.2}
\resizebox{\linewidth}{!}{
\begin{tabular}{l}
\toprule
\textbf{Source sentence:} @GovStitt Please grant clemency for Julius Jones, an innocent man scheduled for execution in your state.\\
\textbf{References:}\\ 
1. @GovStitt Almost like murder if you execute the innocent Julius Jones tomorrow Governor.\\
2. @GovStitt Please commute the sentence of Julius Jones. \\
3. @GovStitt I join the many , many voices urging you to do the right thing and grant clemency to Julius Jones. \\
4. @GovStitt Please save the life of Julius Jones. \\
5. @GovStitt please do the right thing and don't execute julius jones.\\
6. @OKFirstLady Please urge your husband @GovStitt to grant Julius Jones clemency. \\
7. @GovStitt Respectfully I urge you to exercise all powers vested in your office to grant clemency to Mr. Julius Jones \\
8. @GovStitt Please stop the needless execution of Julius Jones! \\ \midrule
\textbf{Source sentence:} Austria imposes COVID-19 lockdown that Applies only to the unvaccinated\\
\textbf{References:}\\ 
1. Austria decided to have a lockdown of the unvaccinated.\\
2. Unvaccinated people forced into lockdown in Austria \\
3. Austria enters hard-to-enforce Covid-19 lockdown for the unvaccinated \\
4. Austria orders non-vaccinated people into COVID-19 lockdown \\
5. Lockdown takes effect for unvaccinated people in Austria \\
6. Unvaccinated People in Austria Are Now Being Put in Lockdown \\
7. Austria orders lockdown for residents who have not received COVID-19 vaccine \\
8. Austria brings back COVID-19 lockdown , this time for the unvaccinated \\ \midrule
\textbf{Source sentence:} Turn off Bluetooth when you are not using it.\\
\textbf{References:}\\ 
1. Reminder to turn off your blue tooth when not in use\\
2. Turn your Bluetooth off while you're not using it. \\
3. Best to turn off Bluetooth when you can. \\
4. Always turn off your Bluetooth when you're not using it \\
5. Whenever you don't absolutely need it, you should go ahead and turn off your Bluetooth. \\
6. Keep Bluetooth off when you are not using it. \\
7. Whenever you don't need BlueTooth, you should turn it off \\
8. If you don't need your Bluetooth enabled, then turn it off! \\
\bottomrule
\end{tabular}}
\caption{Three examples from \newtestdata.}
\label{table:multi_ref_examples}
\end{table*}
\section{Human Evaluation Details}
\label{sec:human_evaluation}
We display our human evaluation instruction for each aspect (fluency, semantic similarity, diversity) in Figure \ref{fig:human_eval_fluency},\ref{fig:human_eval_similarity},\ref{fig:human_eval_diversity}.

\begin{figure*}[t]
    \centering
    \includegraphics[max size={0.95\textwidth}{\textheight}]{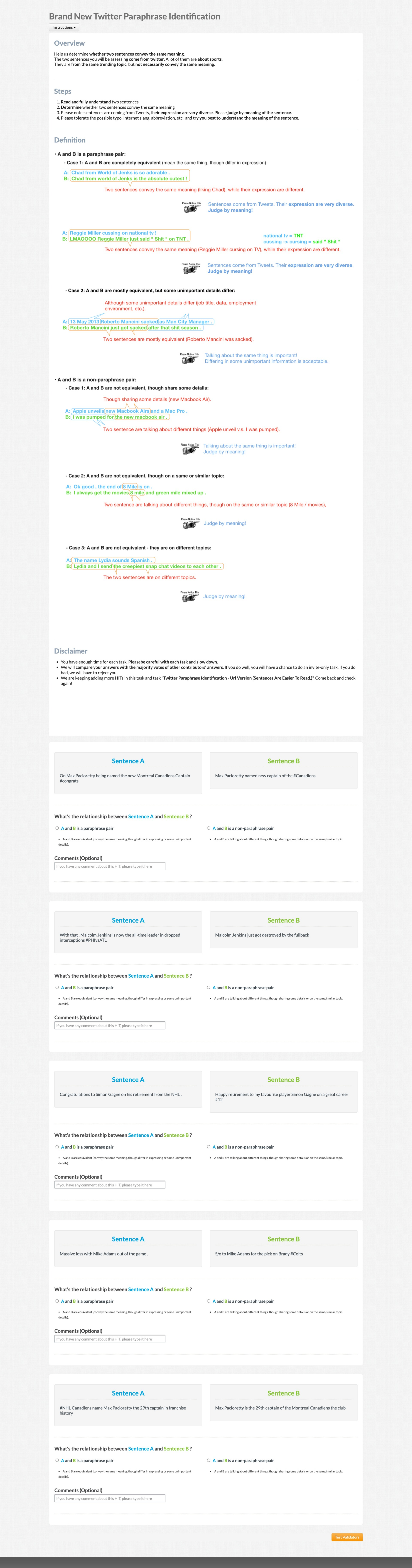}
    \captionof{figure}{Instruction of our crowdsourcing annotation on the Figure Eight platform for creating \crowddata.}
    \label{fig:crowd_annotation_interface}
\end{figure*}

\begin{figure*}[t]
    \centering
    \includegraphics[max size={\textwidth}{\textheight}]{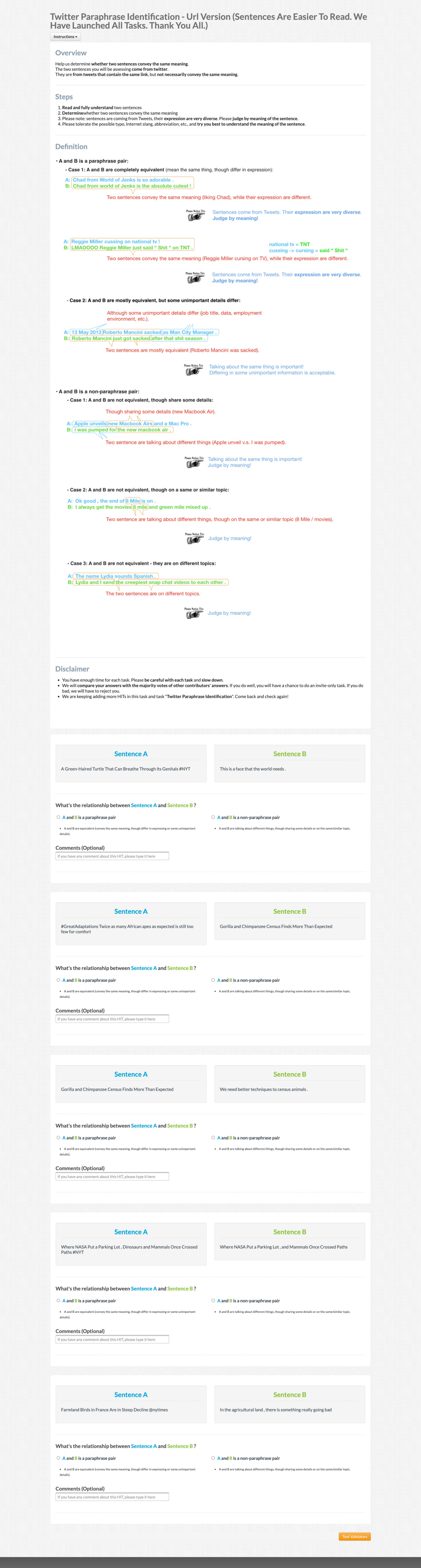}
    \captionof{figure}{An example question of our crowdsourcing annotation on the Figure Eight platform for creating \crowddata.}
    \label{fig:crowd_annotation_question}
\end{figure*}
\begin{figure*}[t]
    \centering
    \includegraphics[width=0.99\textwidth,height=0.95\textheight]{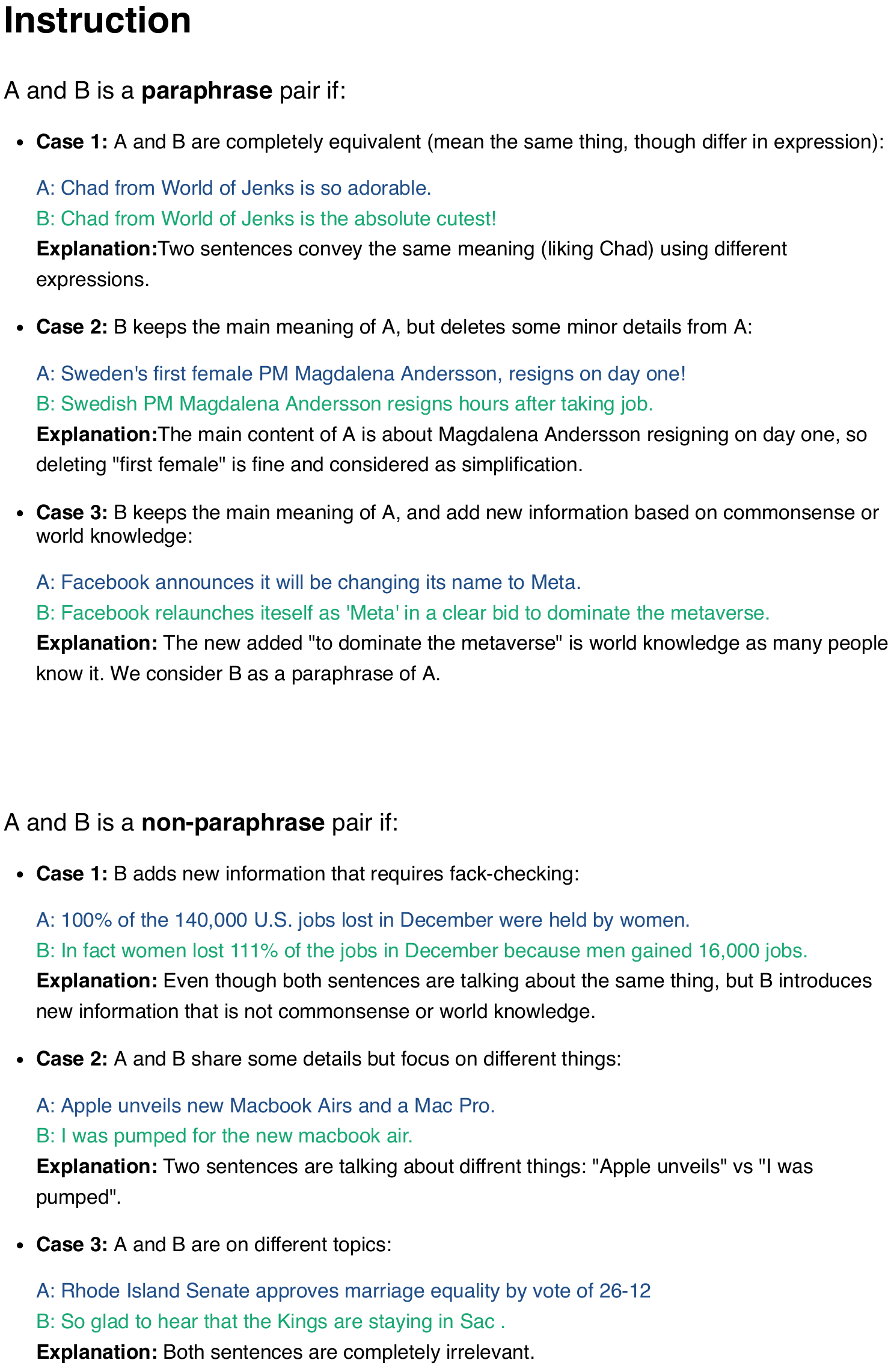}
    \captionof{figure}{Instruction of our expert annotation for creating \expertdata.}
    \label{fig:expert_annotation_interface}
\end{figure*}
\begin{figure*}[t]
    \centering
    \includegraphics[width=0.99\textwidth]{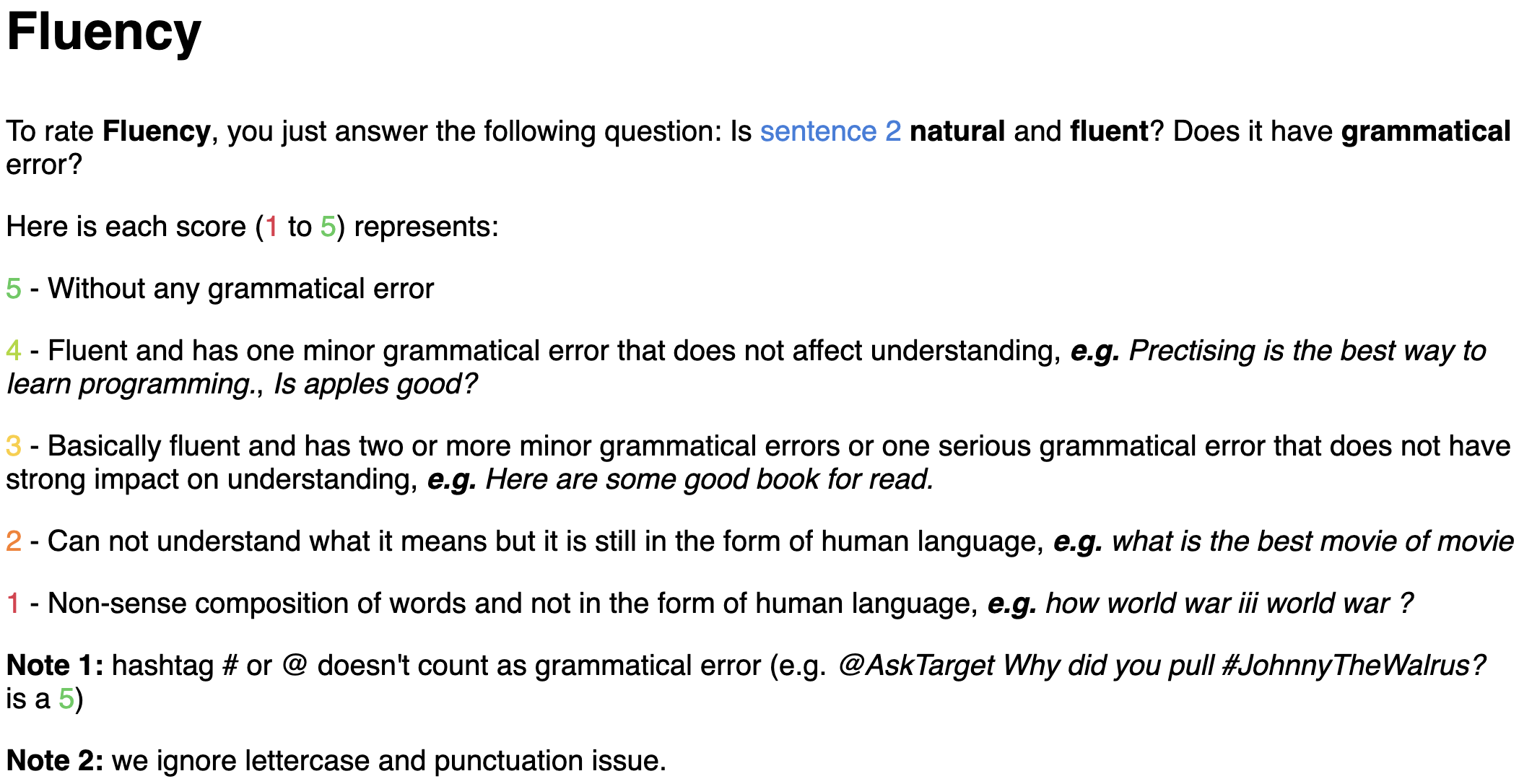}
    \captionof{figure}{Instruction for rating fluency aspect in our human evaluation.}
    \label{fig:human_eval_fluency}
\end{figure*}

\begin{figure*}[t]
    \centering
    \includegraphics[width=0.99\textwidth]{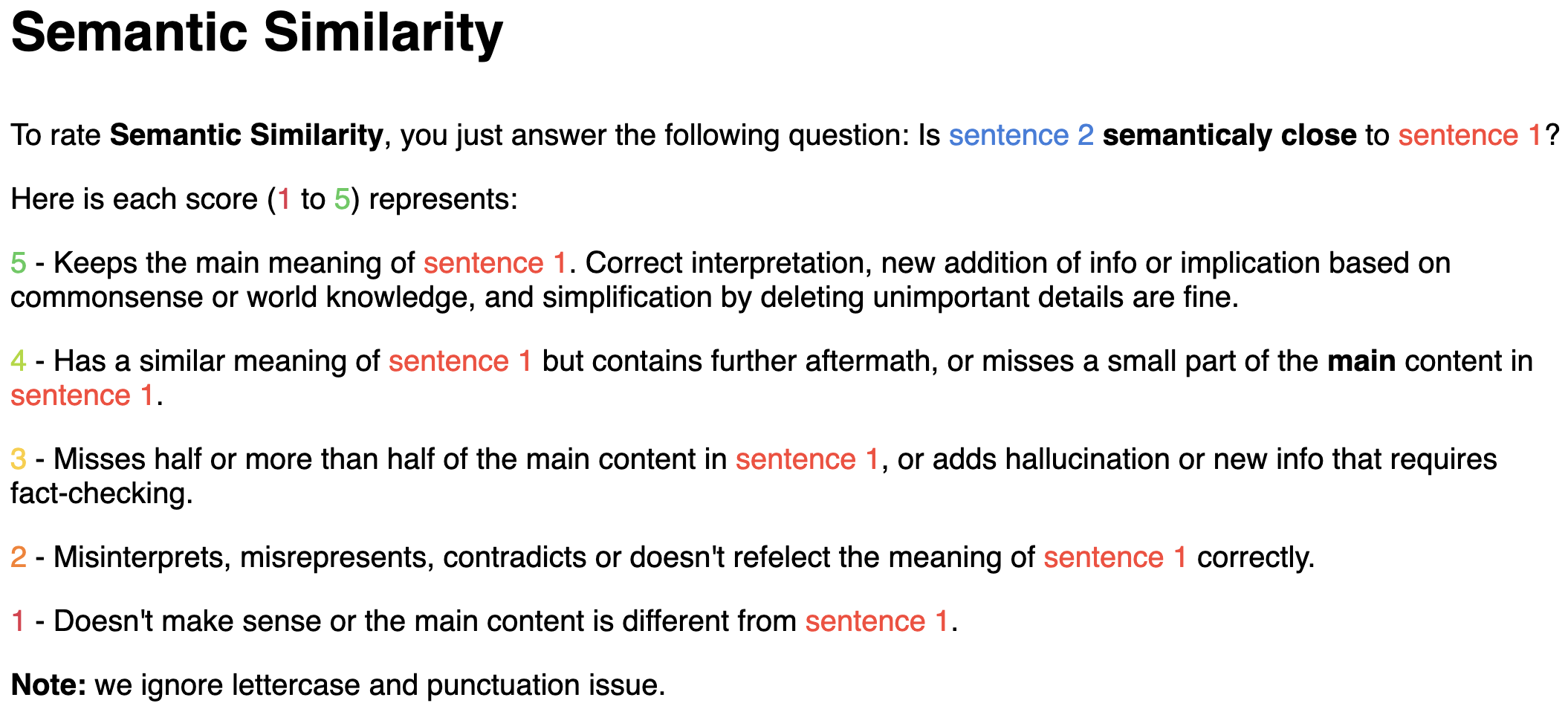}
    \captionof{figure}{Instruction for rating semantic similarity aspect in our human evaluation.}
    \label{fig:human_eval_similarity}
\end{figure*}

\begin{figure*}[t]
    \centering
    \includegraphics[width=0.99\textwidth]{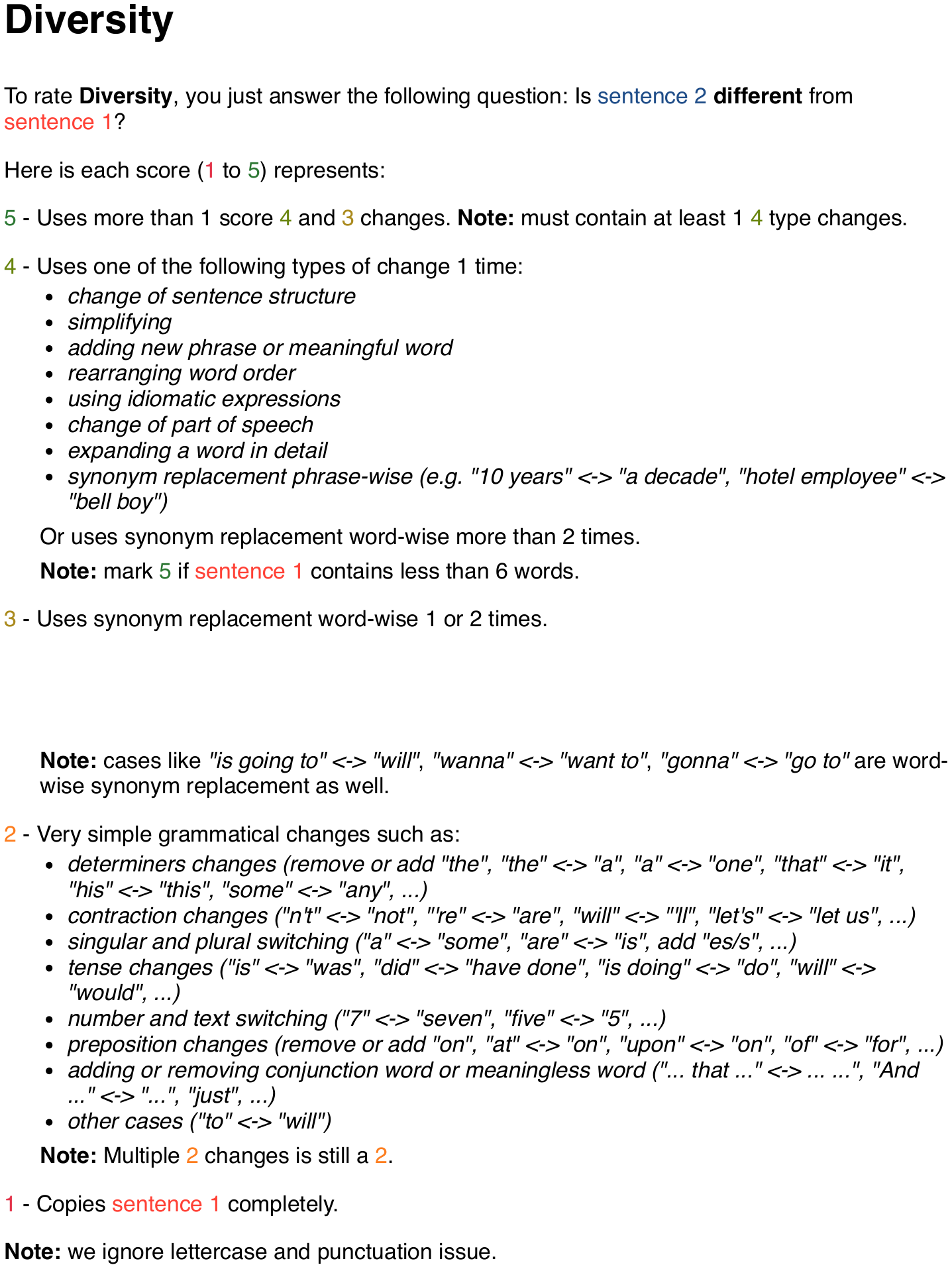}
    \captionof{figure}{Instruction for rating diversity aspect in our human evaluation.}
    \label{fig:human_eval_diversity}
\end{figure*}

\end{document}